\pdfoutput=1

\documentclass[11pt]{article}

\usepackage[final]{acl}

\usepackage{times}
\usepackage{latexsym}

\usepackage[T1]{fontenc}

\usepackage[utf8]{inputenc}

\usepackage{microtype}

\usepackage{inconsolata}

\usepackage{float}
\usepackage{graphicx}
\usepackage{caption}
\usepackage{subcaption}
\usepackage{arydshln}

\title{How Does Knowledge Selection Help Retrieval Augmented Generation?}

\author{Xiangci Li\textsuperscript{\rm 1, \rm 2}$^*$ ~~ Jessica Ouyang\textsuperscript{\rm 2}\\
  \textsuperscript{\rm 1} AWS AI Labs, 
  \textsuperscript{\rm 2} University of Texas at Dallas \\
  {\tt lixiangci8@gmail.com, jessica.ouyang@utdallas.edu} \\
}

\begin{document}
\maketitle
\def\thefootnote{*}\footnotetext{~Work performed outside of AWS AI Labs.}
\begin{abstract}
Retrieval-augmented generation (RAG) is a powerful method for enhancing natural language generation by integrating external knowledge into a model’s output. While prior work has demonstrated the importance of improving knowledge retrieval for boosting generation quality, the role of knowledge \textit{selection}, a.k.a. \textit{reranking} or \textit{filtering}, remains less clear. This paper empirically analyzes how knowledge selection influences downstream generation performance in RAG systems. By simulating different retrieval and selection conditions through a controlled mixture of gold and distractor knowledge, we assess the impact of these factors on generation outcomes. Our findings indicate that the downstream generator model’s capability, as well as the complexity of the task and dataset, significantly influence the impact of knowledge selection on the overall RAG system performance. In typical scenarios, improving the knowledge recall score is key to enhancing generation outcomes, with the knowledge selector providing limited benefit when a strong generator model is used on clear, well-defined tasks. For weaker generator models or more ambiguous tasks and datasets, the knowledge F1 score becomes a critical factor, and the knowledge selector plays a more prominent role in improving overall performance.
\end{abstract}

\section{Introduction}
Retrieval-augmented generation (RAG) has become a pivotal technique in natural language generation, enhancing a language model’s ability to produce relevant, informed output by incorporating external knowledge \cite{gao2023retrieval, fan2024survey, gan2025retrieval}. This approach complements the model’s internal knowledge, which is inherently constrained by the information that was available during training, by retrieving up-to-date, relevant information to support and inform its output.

The quality of the retrieved knowledge directly influences the quality of RAG outputs. Previous studies have consistently shown that improving knowledge \textit{retrieval} leads to a direct enhancement in generation performance \cite{dinan2019wizard, li-etal-2022-enhancing-knowledge, li-etal-2024-knowledge, wang2024retrieve, wu2024how}. Similarly, effective knowledge \textit{selection}, a.k.a. \textit{reranking} or \textit{filtering}, has been observed to improve generation quality by filtering out retrieved information that is less relevant to the generation target \cite{kimsequential, thulkeDSTC2021, li-etal-2022-enhancing-knowledge, sun-etal-2023-generative, zhang2023coarse, wang2023learning, zheng2024ks, wang2024retrieve, zhao-etal-2025-funnelrag}. However, we observe that knowledge selection is barely used for tasks other than dialogue generation, and there are notably fewer LLM-based RAG works that use knowledge selection, compared to fine-tuned RAG models (see Section \ref{sec:rw}). 

We hypothesize that knowledge selectors may not always improve downstream generation performance, and that there may be a selection bias where only positive results involving knowledge selector modules are published, while experiments where knowledge selectors are not helpful simply do not report those results. Further, prior works focus on proposing specific knowledge selection approaches, offering only narrow, case-specific insights into the relationship between knowledge and generation. As a result, readers are often left with anecdotal observations, such as ``using model A in scenario X improves performance'', without a global picture of how knowledge selection impacts generation and when it is  most effective. 

Therefore, in this work, we perform a systematic empirical analysis\footnote{\url{https://github.com/jacklxc/KnowledgeSelectionSimulation}} of how knowledge selectors with various performances affect downstream RAG performance. By blending gold knowledge with distractor knowledge in varying ratios, we \textit{simulate} different selection outcomes and examine their impact on the overall performance of RAG. We find that the generator model’s capability, as well as the complexity of the task and dataset, significantly influence RAG system performance. In typical scenarios, improving knowledge \textit{recall} via the knowledge retriever is key to enhancing generation outcomes, with the knowledge selector providing limited additional benefit when a strong generator model is used on clear, well-defined tasks. For weaker generator models or more ambiguous tasks and datasets, knowledge \textit{F1} becomes a critical factor, and the knowledge selector plays a more prominent role in improving overall performance.

\begin{figure}[t]
\centering
  \includegraphics[width=0.48\textwidth]{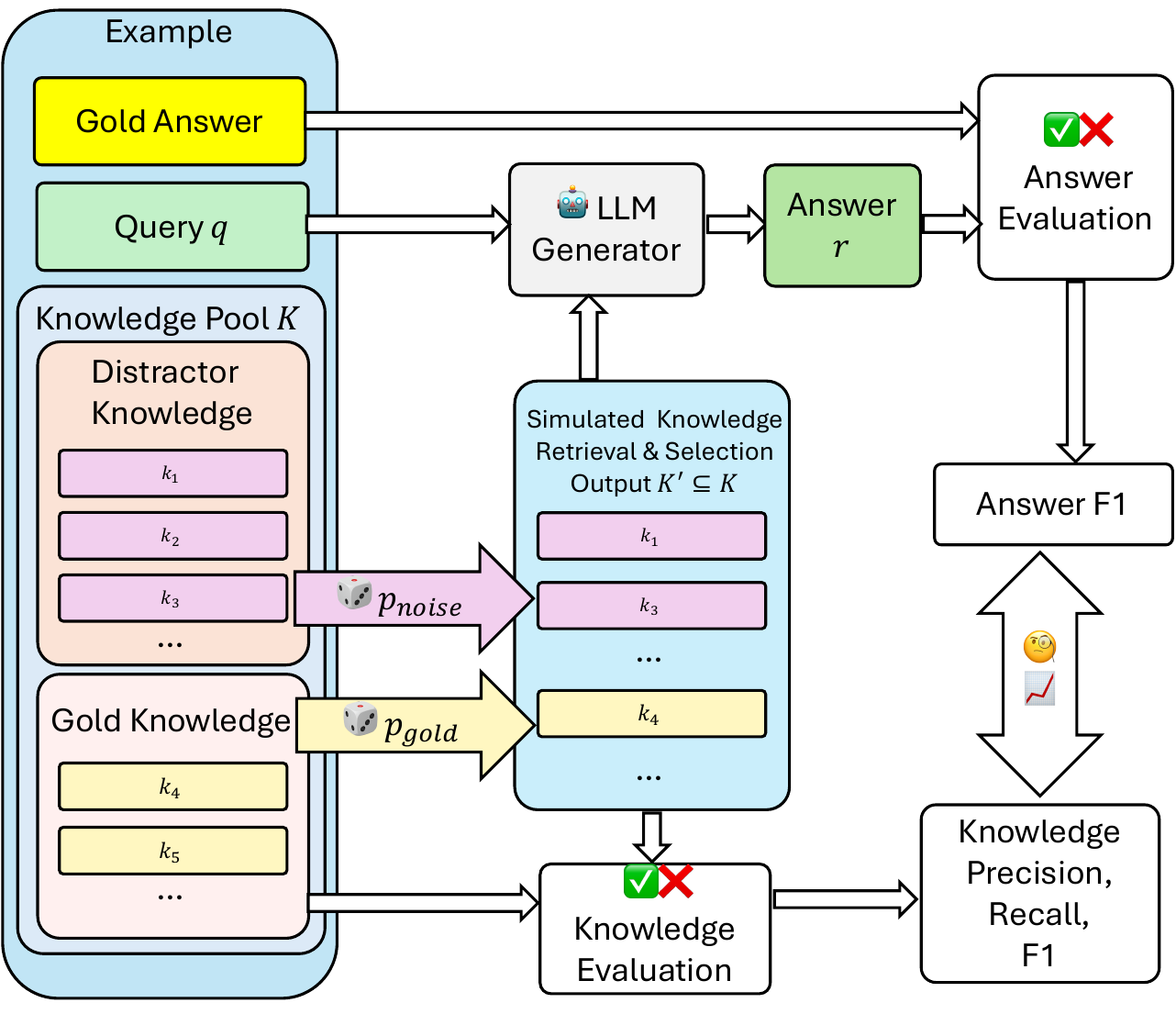}
  \caption{Our simulation experiment pipeline.} 
  \label{fig:rag_simulation_illustration}
\end{figure}

\section{Related Work}
\label{sec:rw}

\paragraph{Retrieval-augmented generation (RAG)} has been extensively studied in recent years. Early works \cite{guu2020retrieval, lewis2020retrieval, shuster-etal-2021-retrieval-augmentation} jointly fine-tuned a dense retriever (e.g. DPR \cite{karpukhin-etal-2020-dense}) and generator (e.g. BART \cite{lewis-etal-2020-bart}), which required dedicated training datasets. More recently, the introduction of large language models (LLMs) has made RAG more convenient to implement due to their strong generation performance, in-context learning ability \cite{brown2020language, kojima2022large}, and drastically longer context windows --- from BART's 1024 tokens to more than a million for modern LLMs \cite{lee2024can}. As a result, recent RAG research has shifted to using LLMs \cite{gao2023retrieval, fan2024survey, gan2025retrieval}; in this work, we follow this trend to focus on LLM-based RAG.

\paragraph{Knowledge selector for knowledge-grounded dialogue generation.} Dialogue generation \cite{moghe-etal-2018-towards, dinan2019wizard, li-etal-2024-knowledge} is one of the major applications of RAG, where the target response is conditioned on retrieved knowledge, and a knowledge selection step is often added to further refine the retrieved knowledge \cite{thulkeDSTC2021, sun-etal-2023-generative, zhang2023coarse}. For example, \citet{kimsequential} train a knowledge selector by leveraging response information, 
\citet{li-etal-2022-enhancing-knowledge} select knowledge from document semantic graphs, \citet{zhang2023coarse} propose multi-task learning for knowledge selection and response generation, and \citet{zhao-etal-2025-funnelrag} proposes a multi-step reranking process for Natural Question \cite{kwiatkowski-etal-2019-natural} and Trivia QA \cite{joshi-etal-2017-triviaqa}. However, while these works demonstrate the advantages of their approaches through ablation studies, it is unclear whether their performance improvement via knowledge selection specifically is transferable to other datasets, domains, or tasks. 
Moreover, despite the popularity of LLMs for RAG, there are dramatically fewer LLM-based works that include the knowledge selection step, and it is likewise omitted in other RAG tasks \cite{gao2023retrieval, fan2024survey, gan2025retrieval}. 

To our knowledge, there is no prior work explaining this gap in knowledge \textit{selection}. The closest works to ours are \citet{cuconasu2024power}, \citet{wu2024how} and \citet{jin2025longcontext}, which study the impact of knowledge \textit{retrieval} on RAG. We hypothesize that there is a selection bias effect where knowledge selection modules may have been experimented with, but ultimately not reported, in tasks or settings where it is not helpful. In this paper, we study the general effect of knowledge retrieval and selection on LLM-based RAG via hundreds of simulations, rather than being restricted to a few specific, anecdotal retrievers or selector implementations like prior works.

\section{Approach}

\subsection{Task Formulation}
RAG involves three steps: (1) Knowledge retrieval, where a retriever module retrieves a set of candidate knowledge $K$ based on a query $q$. This step aims to retrieve as much knowledge that is relevant to the query as possible, balancing knowledge \textit{recall} and \textit{precision}. (2) Optionally, knowledge selection, a.k.a. reranking or filtering, removes retrieved knowledge that is less relevant to further improve knowledge \textit{precision}, producing $K' \subseteq K$. (3) Finally, the generator takes the query $q$ and the selected, retrieved knowledge $K'$ to generate the output text $r$. 

The knowledge in $K$ and $K'$ can be heterogeneous and come from multiple sources, such as external documents, knowledge graphs, or conversation histories. This RAG framework can be applied to various tasks, such as dialogue generation, question answering, fact verification, and code generation \cite{gao2023retrieval, fan2024survey, gan2025retrieval}. In our experiments on dialogue generation and question answering, we \textit{simulate} steps 1 and 2 by creating controlled knowledge sets $K'$ and observing the resulting generation performance of step 3.

\subsection{Knowledge Simulation} \label{sec:simulation_approach}
We aim to perform a systematic analysis of the effect of knowledge retrieval and selection outcomes on the downstream RAG performance by \textit{simulation}. 
As Figure \ref{fig:rag_simulation_illustration} shows, for each query $q$, given a fixed pool of available knowledge with gold relevance annotations, we sample gold knowledge and distractor knowledge at varying rates $p_{gold}$ and $p_{noise}$ to precisely simulate a wide range of quality for the retrieved and selected knowledge $K'$, from noise only to gold knowledge only. For example, if $p_{gold}=0.5$, then each piece of gold knowledge has a 50\% chance to be sampled. Each sampling produces a full experiment over the \textit{entire test set}, reported as one data point in Figures \ref{fig:selection_prec_recall_vs_answer_f1}-\ref{fig:fixed_recall_precision_vs_answer_f1}. 

In this way, we simulate a wide distribution of knowledge retriever and selector performance, as measured by knowledge precision and recall based on the gold annotations. We then test the performance of the generator model given the different quality of simulated retrieved and selected knowledge $K'$. Compared to prior works discussed in Section \ref{sec:rw}, which only compare a few ablation configurations, we conduct hundreds of configurations in each meta-experiment.

\section{Experimental Settings}
\subsection{Datasets} \label{sec:datasets}
While there are several datasets supporting the RAG framework, relatively few provide high-quality, human-annotated gold knowledge. Furthermore, the target outputs should be relatively short, unambiguous, and easily evaluated with automatic metrics such as F1 scores. These conditions narrow down the choices to two popular and representative datasets, Wizard of Wikipedia \citep[WoW;][]{dinan2019wizard} and HotpotQA \cite{yang-etal-2018-hotpotqa}.

\paragraph{WoW} is an open-domain dialogue dataset based on Wikipedia knowledge that has been widely used in prior knowledge selection works \cite{kimsequential, thulkeDSTC2021, li-etal-2022-enhancing-knowledge, sun-etal-2023-generative}. The wizard speaker (a human annotator playing the part of the dialogue system) has access to Wikipedia knowledge, while the apprentice speaker does not. The apprentice is given a starting conversation topic, but otherwise speaks freely. The last two turns of dialogue are used as the query $q$ to \textit{retrieve} Wikipedia passages ($K$), and the wizard \textit{selects} a single knowledge sentence ($K'$) to generate their response. 

Despite its popularity, this dialogue generation task is challenging to evaluate. First, because the wizard is limited to selecting only a single knowledge sentence, 
it is still possible for the unselected ``distractor" knowledge to be relevant to the wizard's response. Second, as is the case in many natural language generation tasks, the gold wizard responses in WoW are not the only plausible responses, making it harder to quantify the correctness of a generated response.
Nonetheless, WoW is a well-annotated dialogue dataset that is close to a real-life scenario where the gold knowledge and responses are noisy.

\paragraph{HotpotQA} is a question-answering dataset derived from Wikipedia knowledge that we include in our experiments to mitigate the ambiguity of WoW. The multi-hop questions and answers are first directly derived from gold knowledge graphs, and then distractor knowledge is injected to create noise. As a result, unlike WoW, the answers in HotpotQA are strongly dependent on the gold knowledge, and the distractor knowledge is unlikely to be relevant. The short and unambiguous gold answers make it simpler to evaluate via F1 scores.


\subsection{Generators}
To align with the latest trends in RAG research, as well as to make the analysis simple and generally replicable, we adopt LLMs as our generators. Due to the large computational costs for the meta-experiments, each of which consists of hundreds of full experiments, we use three API-based lightweight LLMs, OpenAI GPT-4o-mini, LLaMA 3.1 8B, and Mistral 7B-Instruct, as our generator models; the varying performances of the LLMs allow us to investigate the impact of generator complexity on knowledge usage.

\begin{table}[t]
\setlength{\tabcolsep}{2pt} 
\begin{center}
\small
    \begin{tabular}{cccccl}
    \hline
    \textbf{Input Knowledge} & \textbf{KP} & \textbf{KR} & \textbf{KF1} & \textbf{R-L} & \textbf{F1} \\ \hline
     \multicolumn{6}{l}{\textit{GPT-4o-mini}} \\
    No knowledge & 0 & 0 & 0 & 0.110 & 0.200 ($\pm$ .005)\\
    Full knowledge & 0.015 & 1 & 0.031 & 0.140 & 0.251 ($\pm$ .006) \\
    Gold knowledge & 1 & 1 & 1 & 0.167 & 0.276 ($\pm$ .007) \\ \hline 
    \multicolumn{6}{l}{\textit{LLaMA 3.1 8B}} \\ 
    No knowledge & 0 & 0 & 0 & 0.111 & 0.216 ($\pm$ 0.005) \\
    Full knowledge & 0.015 & 1 & 0.031 & 0.138 & 0.248 ($\pm$ .005) \\
    Gold knowledge & 1 & 1 & 1 & 0.164 & 0.278 ($\pm$ .008) \\ \hline 
    \multicolumn{6}{l}{\textit{Mistral 7B Instruct}} \\ 
    No knowledge & 0 & 0 & 0 & 0.113 & 0.203 ($\pm$ .005) \\
    Full knowledge & 0 & 1 & 0 & 0.131 & 0.233 ($\pm$ .005) \\
    Gold knowledge & 1 & 1 & 1 & 0.172 & 0.268 ($\pm$ .007) \\ \hline 

    \end{tabular}
    \caption{WoW response generation performance benchmarked by different LLM generators. We measure knowledge precision (KP), recall (KR), and F1 (KF1); response ROUGE-L F1 (R-L); and response F1 (and its standard error mean).} \label{tab:wow_benchmark}
\end{center}
\end{table}

\begin{table}[t]
\setlength{\tabcolsep}{2pt} 
\begin{center}
\small
    \begin{tabular}{cccccl}
    \hline
    \textbf{Input Knowledge} & \textbf{KP} & \textbf{KR} & \textbf{KF1} & \textbf{EM} & \textbf{F1} \\ \hline
     \multicolumn{6}{l}{\textit{GPT-4o-mini}} \\
    No knowledge & 0 & 0 & 0 & 0.330 & 0.437 ($\pm$ .020) \\
    Full knowledge & 0.065 & 1 & 0.120 & 0.668 & 0.780 ($\pm$ .016) \\
    Gold knowledge & 1 & 1 & 1 & 0.710 & 0.828 ($\pm$ .014) \\ \hline 
    \multicolumn{6}{l}{\textit{LLaMA 3.1 8B}} \\ 
    No knowledge & 0 & 0 & 0 & 0.200 & 0.298 ($\pm$ .019) \\
    Full knowledge & 0.065 & 1 & 0.120 & 0.372 & 0.545 ($\pm$ .019) \\
    Gold knowledge & 1 & 1 & 1 & 0.414 & 0.671 ($\pm$ .016) \\ \hline 
    \multicolumn{6}{l}{\textit{Mistral 7B Instruct}} \\ 
    No knowledge & 0 & 0 & 0 & 0.208 & 0.260 ($\pm$ .019) \\
    Full knowledge & 0.065 & 1 & 0.120 & 0.046 & 0.151 ($\pm$ .011) \\
    Gold knowledge & 1 & 1 & 1 & 0.502 & 0.627 ($\pm$ .019) \\ \hline 

    \end{tabular}
    \caption{HotpotQA answer generation performance benchmarked by different LLM generators. We measure knowledge precision (KP), recall (KR), and F1 (KF1); answer exact match (EM); and answer F1 (and its standard error mean).} \label{tab:hotpot_qa_benchmark}
\end{center}
\end{table}

\subsection{Knowledge Sampling}

As Figure \ref{fig:rag_simulation_illustration} shows, both WoW and HotpotQA are \textit{knowledge selector training datasets} that provide a retrieved knowledge set $K$ for each query $q$ such that the gold knowledge is a subset of $K$. In both datasets, $K$ 
contains text passages consisting of an article title and a few knowledge sentences. We perform knowledge sampling at the sentence level to simulate the end result of applying both the knowledge retrieval and selection steps. More details are in Appendix \ref{sec:sampling_detail}.

In our experiments, we refer to using the entire retrieved knowledge set $K$ provided by the dataset as the input knowledge to the generator (i.e. no knowledge selection) as the ``full knowledge'' setting, where both gold and distractor knowledge are present. We also include a ``no knowledge" setting, where the generators receive no external knowledge at all, as a weak baseline, and ``gold knowledge," corresponding to perfect knowledge selection, as an upper bound. 

\begin{figure*}[t]
     \centering
        \begin{subfigure}[b]{0.42\textwidth}
        \centering
          \includegraphics[width=\textwidth]{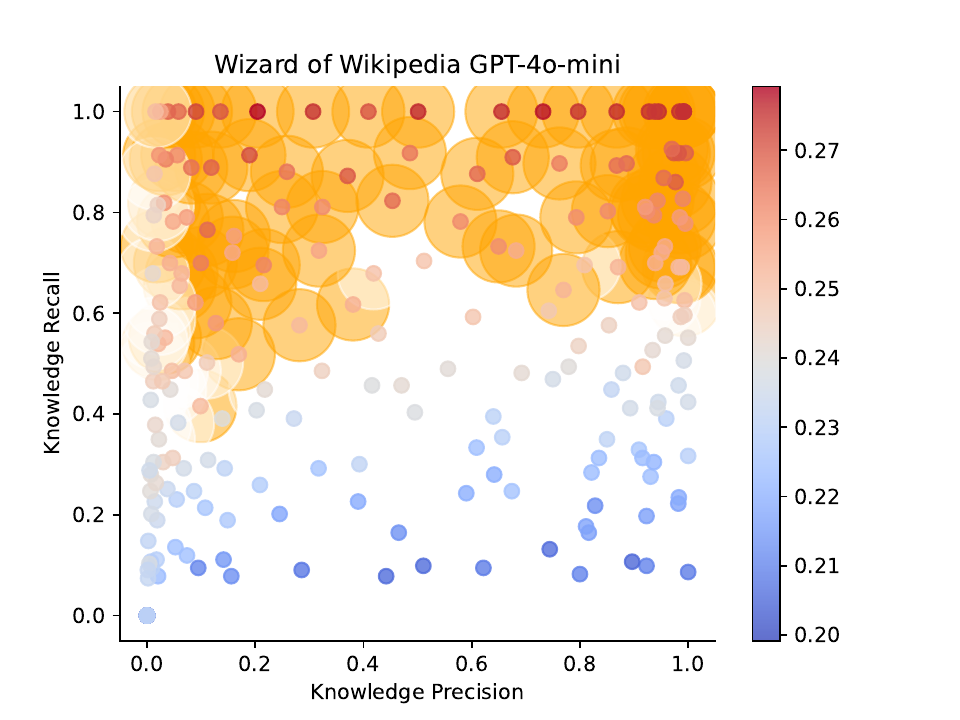}
          \label{fig:wow_gpt_selection_prec_recall_vs_answer_f1}
          \vspace{-1.5em}
        \end{subfigure}
    \quad
        \begin{subfigure}[b]{0.42\textwidth}
        \centering
          \includegraphics[width=\textwidth]{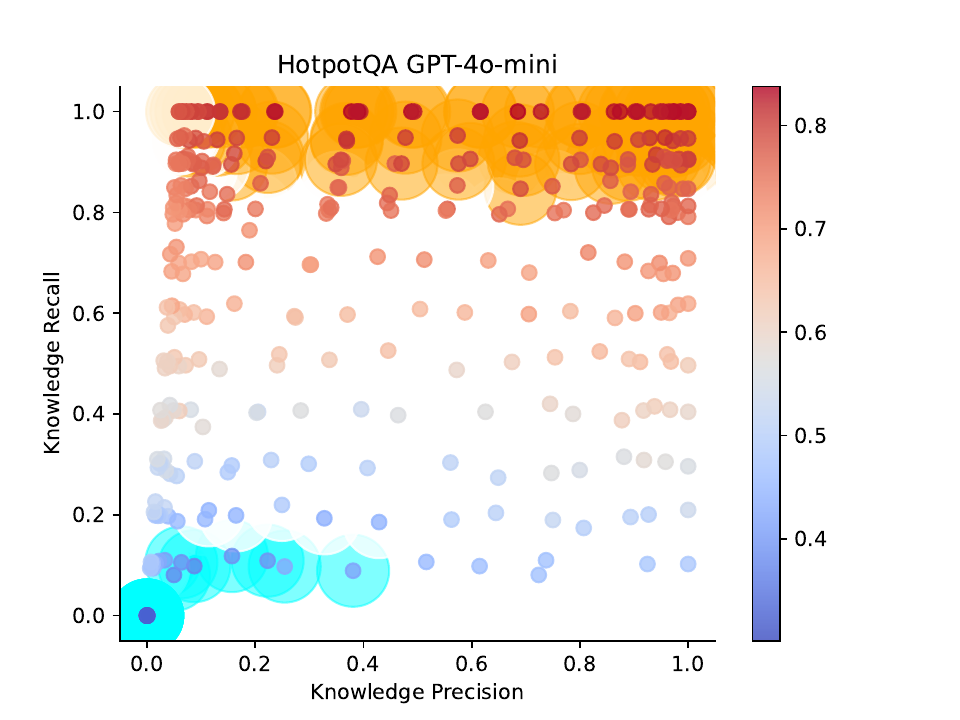}
          \label{fig:hotpot_qa_gpt_selection_prec_recall_vs_answer_f1}
          \vspace{-1.5em}
        \end{subfigure}
     \quad
        \begin{subfigure}[b]{0.42\textwidth}
        \centering
          \includegraphics[width=\textwidth]{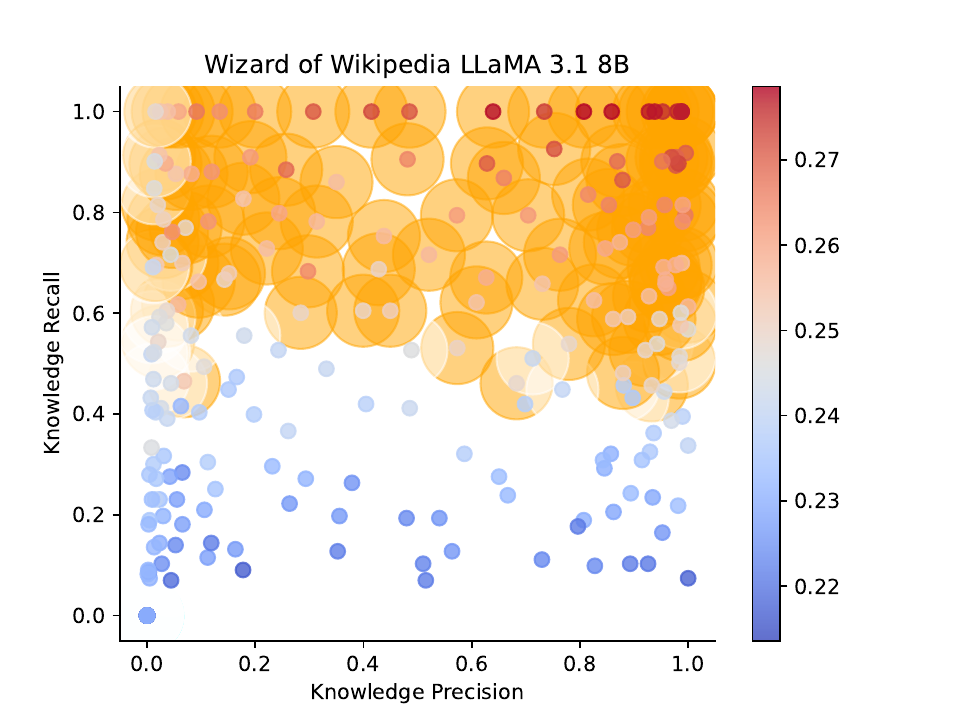}
          \label{fig:wow_llama_selection_prec_recall_vs_answer_f1}
          \vspace{-1.5em}
        \end{subfigure}
    \quad
        \begin{subfigure}[b]{0.42\textwidth}
        \centering
          \includegraphics[width=\textwidth]{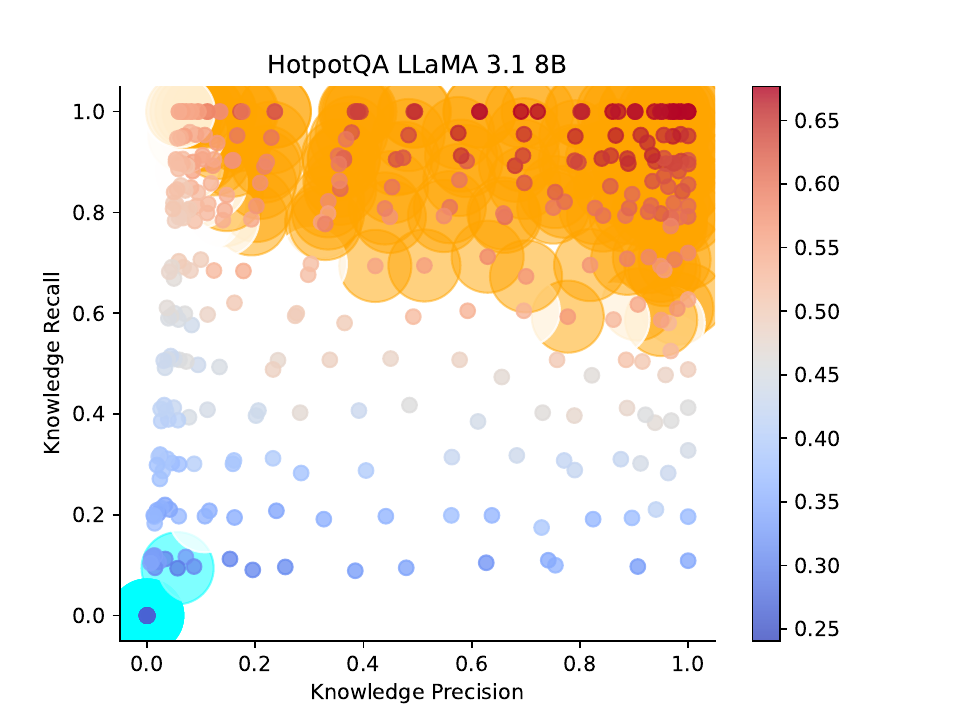}
          \label{fig:hotpot_qa_llama_selection_prec_recall_vs_answer_f1}
          \vspace{-1.5em}
        \end{subfigure}
     \quad
        \begin{subfigure}[b]{0.42\textwidth}
        \centering
          \includegraphics[width=\textwidth]{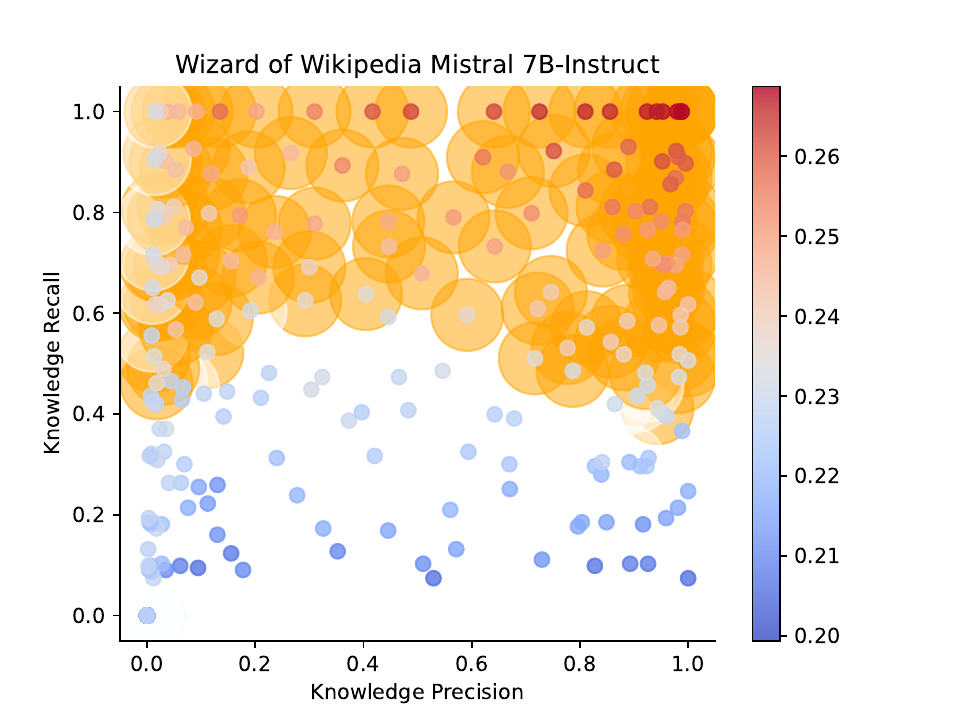}
          \label{fig:wow_mistral_selection_prec_recall_vs_answer_f1}
          \vspace{-1.5em}
        \end{subfigure}
     \quad
        \begin{subfigure}[b]{0.42\textwidth}
        \centering
          \includegraphics[width=\textwidth]{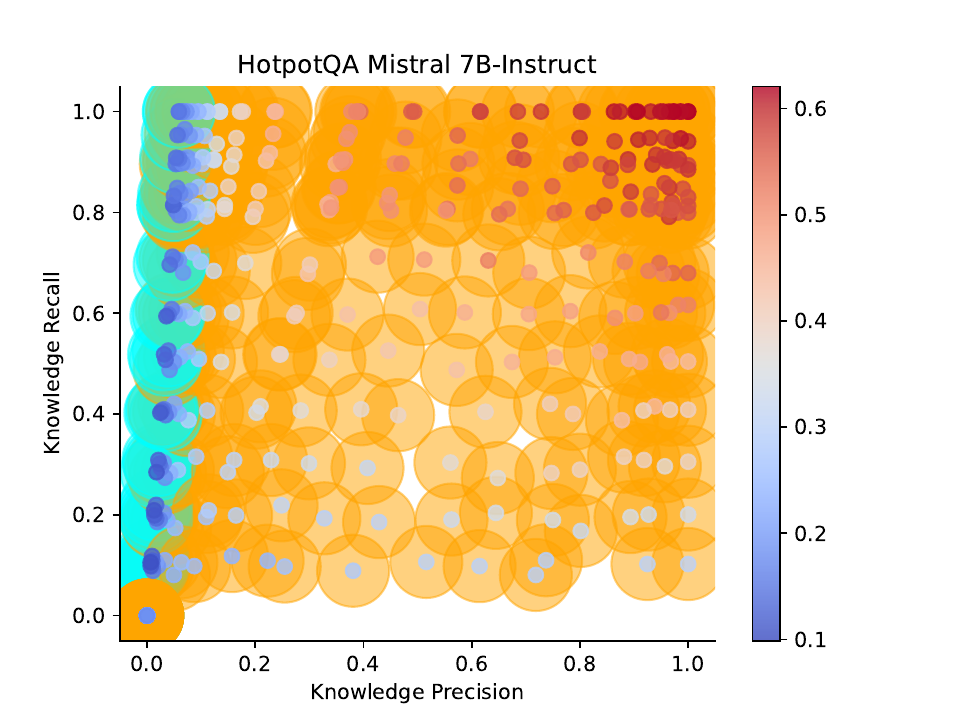}
          \label{fig:hotpot_qa_mistral_selection_prec_recall_vs_answer_f1}
          \vspace{-1.5em}
        \end{subfigure}
     \quad
     \vspace{-1.0em}
        \caption{Scatter plot of response/answer F1, plotted against knowledge precision (x-axis) and recall (y-axis), by GPT-4o-mini (top), LLaMA 3.1 8B (middle), and Mistral 7B-Instruct (bottom). The left column shows results on WoW; the right shows HotpotQA. The dots highlighted in orange indicate settings outperforming the ``full knowledge'' setting, while those highlighted in cyan indicate settings underperforming the ``no knowledge" setting. Each figure is a meta-experiment, and each data point corresponds to a full experiment on the entire sampled dataset.}
        \label{fig:selection_prec_recall_vs_answer_f1}
      \vspace{-1.5em}
\end{figure*}

\section{Meta-Experimental Results}
\label{sec:RagSimulation_results}
\begin{figure*}[t]
     \centering
        \begin{subfigure}[b]{0.42\textwidth}
        \centering
          \includegraphics[width=\textwidth]{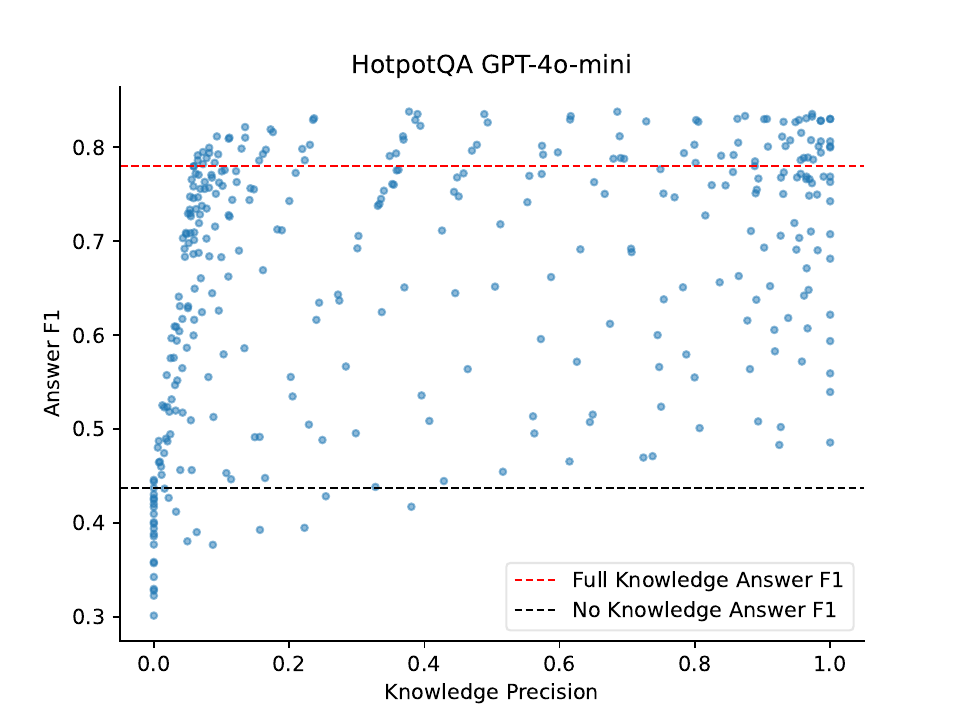}
          \label{fig:hotpot_qa_gpt_selection_prec_vs_answer_f1}
          \vspace{-1.5em}
        \end{subfigure}
        \quad
      \begin{subfigure}[b]{0.42\textwidth}
        \centering
          \includegraphics[width=\textwidth]{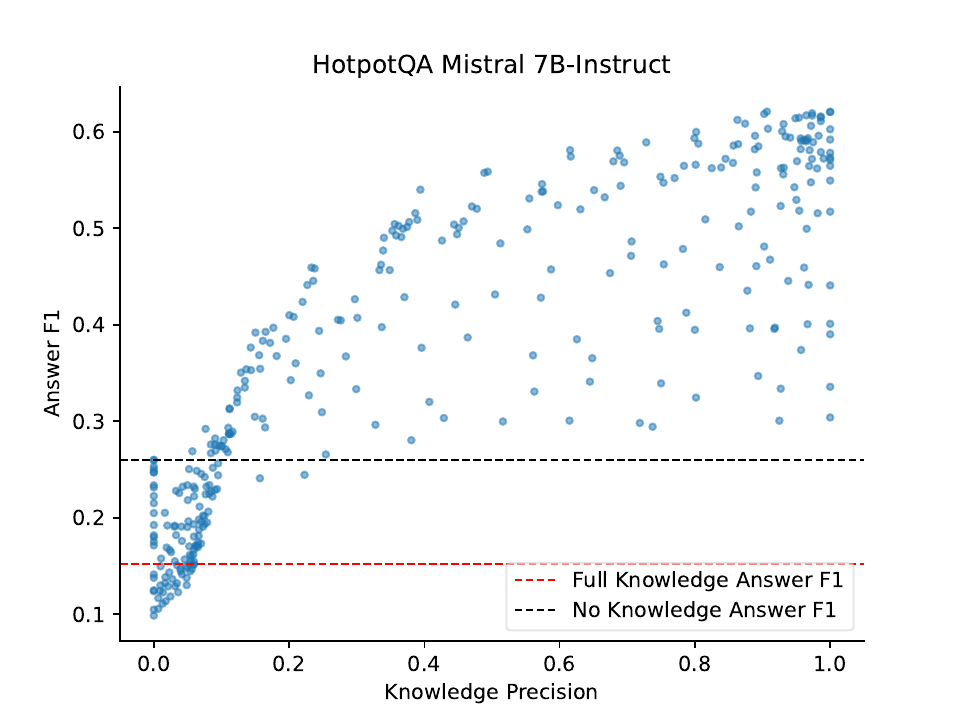}
          \label{fig:hotpotqa_mistral_selection_prec_vs_answer_f1}
          \vspace{-1.5em}
        \end{subfigure}
     \quad
        \begin{subfigure}[b]{0.42\textwidth}
        \centering
          \includegraphics[width=\textwidth]{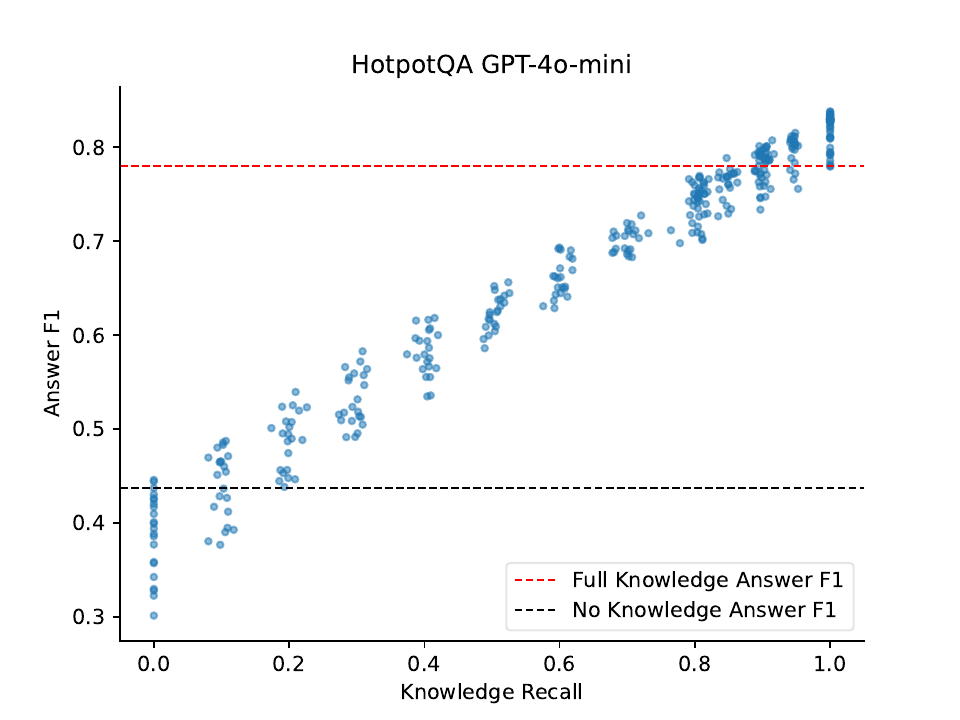}
          \label{fig:hotpot_qa_gpt_selection_recall_vs_answer_f1}
          \vspace{-1.5em}
        \end{subfigure}
     \quad
        \begin{subfigure}[b]{0.42\textwidth}
        \centering
          \includegraphics[width=\textwidth]{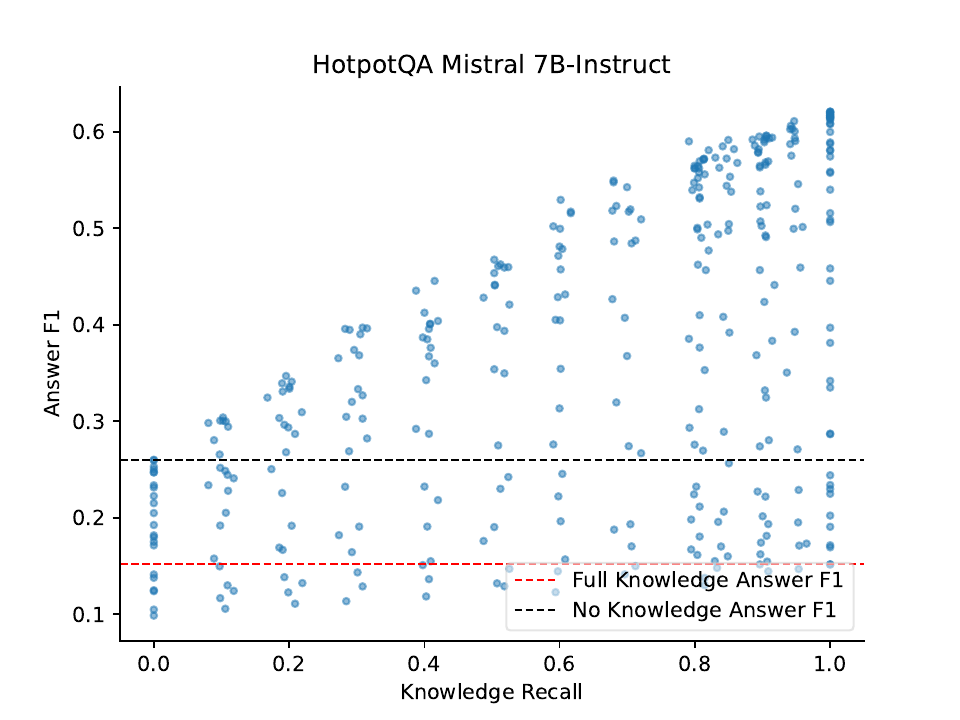}
          \label{fig:hotpotqa_mistral_selection_recall_vs_answer_f1}
          \vspace{-1.5em}
        \end{subfigure}
     \quad
        \begin{subfigure}[b]{0.42\textwidth}
        \centering
          \includegraphics[width=\textwidth]{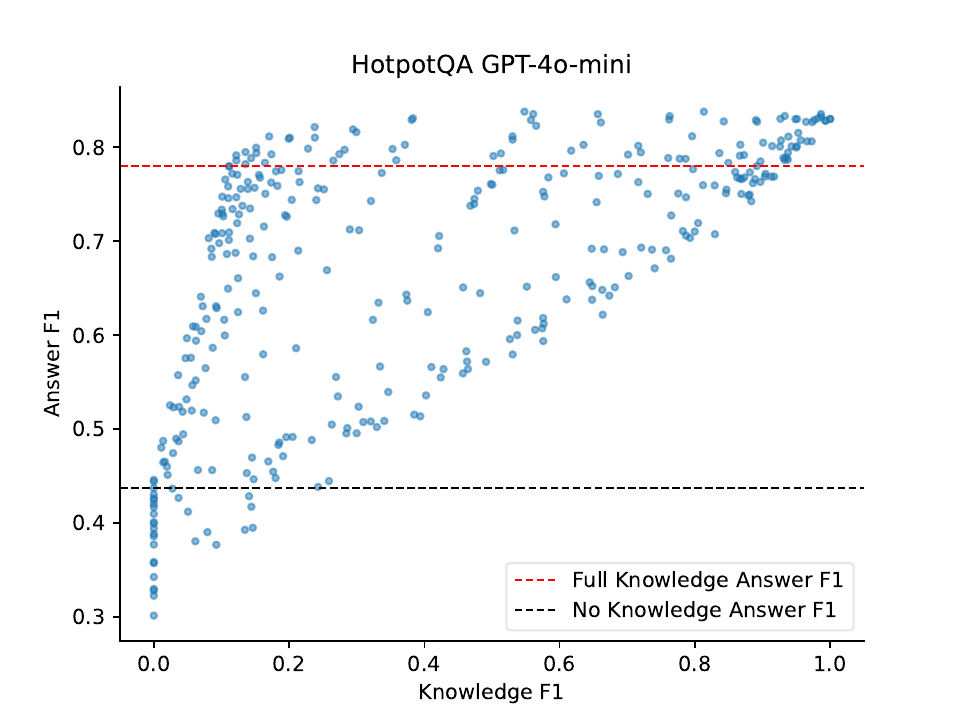}
          \label{fig:hotpot_qa_selection_f1_vs_answer_f1}
          \vspace{-1.5em}
        \end{subfigure}
     \quad
        \begin{subfigure}[b]{0.42\textwidth}
        \centering
          \includegraphics[width=\textwidth]{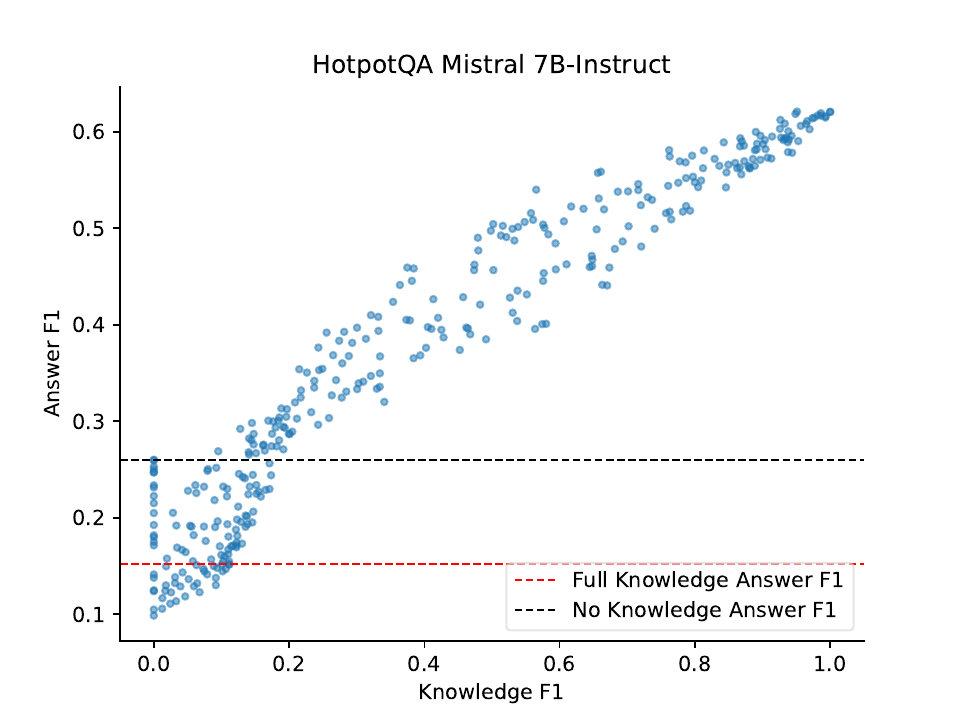}
          \label{fig:hotpotqa_mistral_selection_f1_vs_answer_f1}
          \vspace{-1.5em}
        \end{subfigure}
     \quad
     \vspace{-0.5em}
        \caption{Scatter plot of HotpotQA answer F1 versus knowledge precision (top), knowledge recall (middle), and knowledge F1 (bottom). The left column shows GPT-4o-mini as the generator; the right column shows Mistral-7B-Instruct. Plots for LLaMa 3.1 8B and the WoW dataset are in Appendix \ref{sec:appendix}. Each figure is a meta-experiment, and each data point corresponds to a full experiment on the entire sampled dataset.}
        \label{fig:hotpotqa_selection_vs_answer_f1}
     \vspace{-1.5em}
\end{figure*}

\begin{figure*}[t]
\centering
    \begin{subfigure}[b]{0.4\textwidth}
     \centering
     \includegraphics[width=\textwidth]{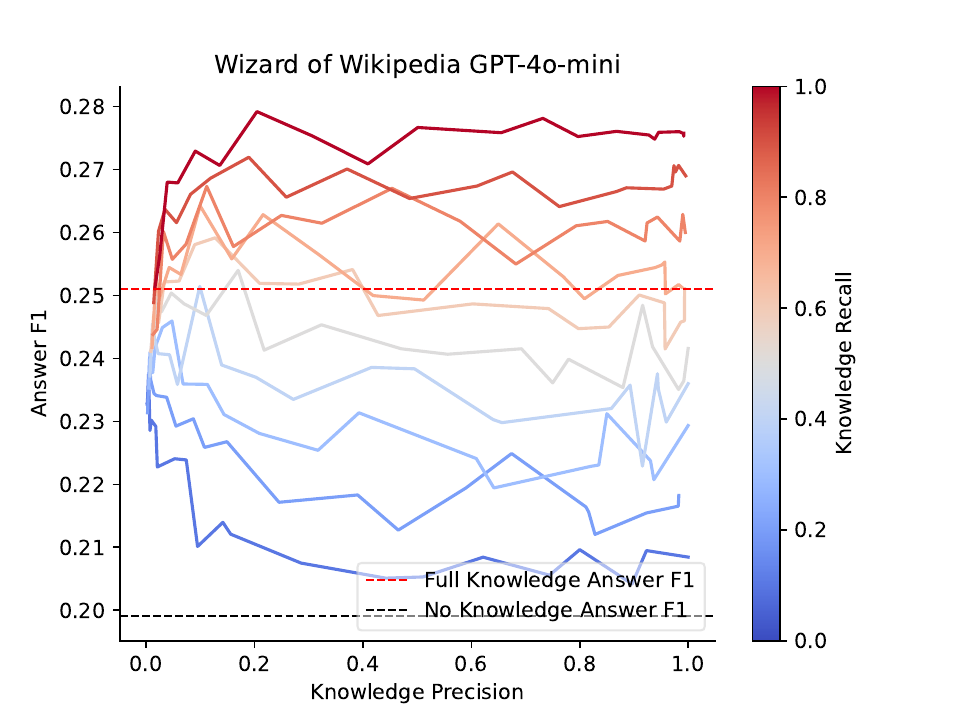}
    \label{fig:wow_gpu_fixed_recall_precision_vs_answer_f1}
    \vspace{-1.5em} 
    \end{subfigure}
    \quad
    \begin{subfigure}[b]{0.4\textwidth}
     \centering
     \includegraphics[width=\textwidth]{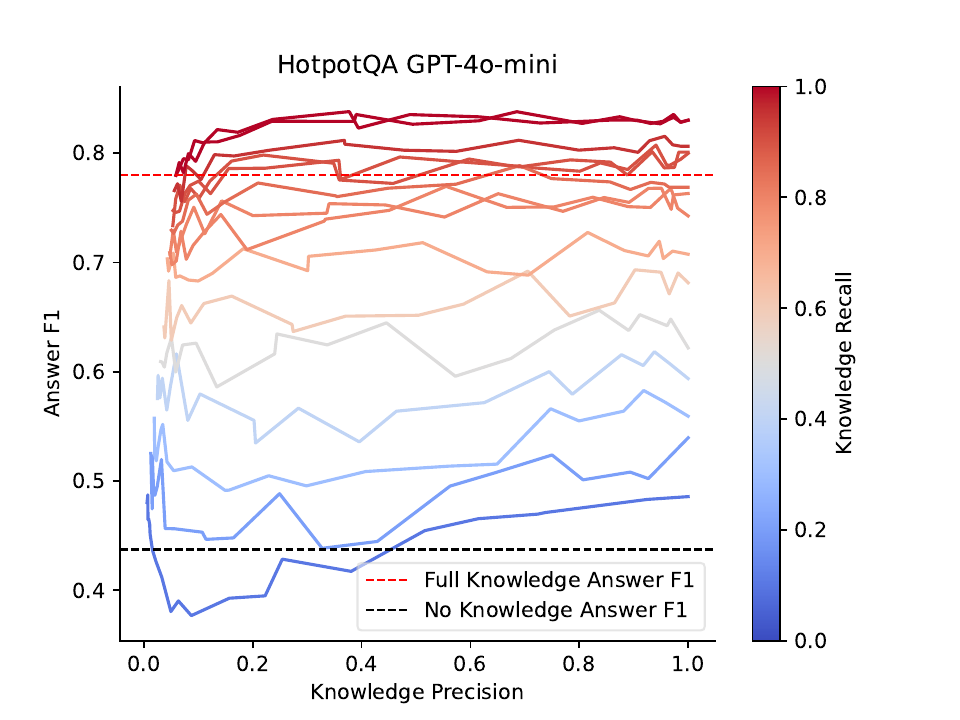}
    \label{fig:hotpot_qa_gpt_fixed_recall_precision_vs_answer_f1}
    \vspace{-1.5em} 
    \end{subfigure}
    \quad
    \begin{subfigure}[b]{0.4\textwidth}
     \centering
     \includegraphics[width=\textwidth]{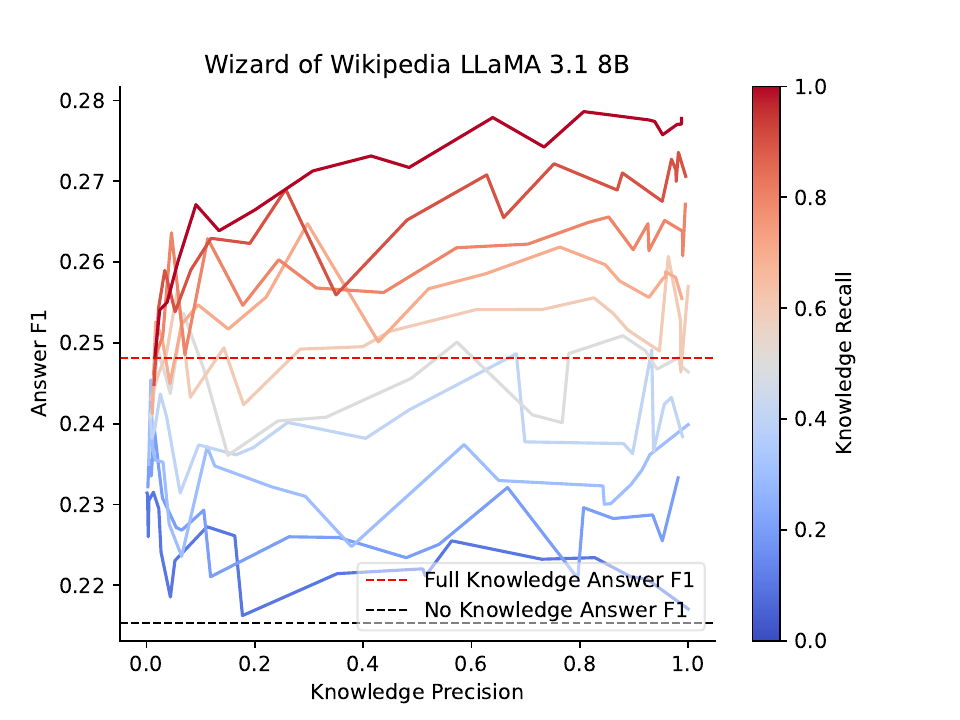}
    \label{fig:wow_llama_fixed_recall_precision_vs_answer_f1}
    \vspace{-1.5em} 
    \end{subfigure}
    \quad
    \begin{subfigure}[b]{0.4\textwidth}
     \centering
     \includegraphics[width=\textwidth]{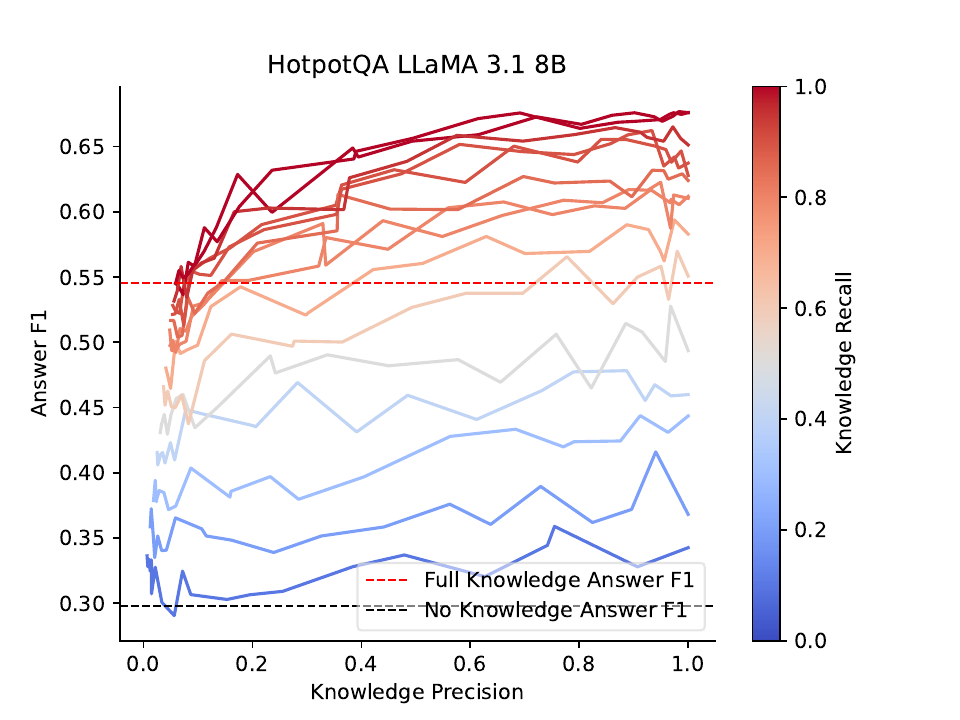}
    \label{fig:hotpot_qa_llama_fixed_recall_precision_vs_answer_f1}
    \vspace{-1.5em} 
    \end{subfigure}
    \quad
    \begin{subfigure}[b]{0.4\textwidth}
     \centering
     \includegraphics[width=\textwidth]{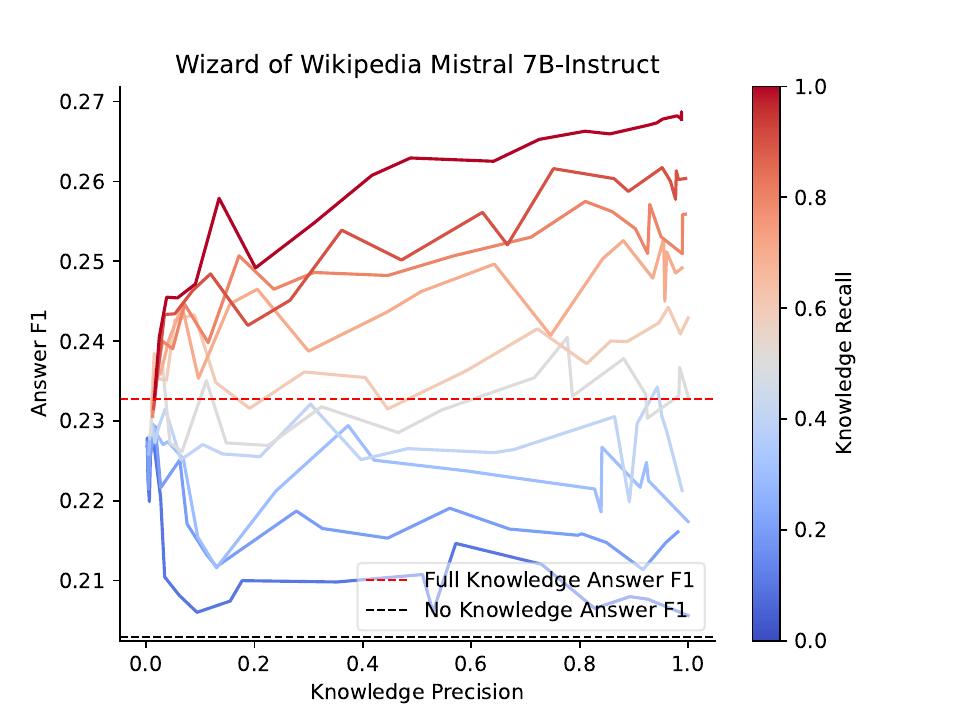}
    \label{fig:wow_mistral_fixed_recall_precision_vs_answer_f1}
    \vspace{-1.5em} 
    \end{subfigure}
    \quad
    \begin{subfigure}[b]{0.4\textwidth}
     \centering
     \includegraphics[width=\textwidth]{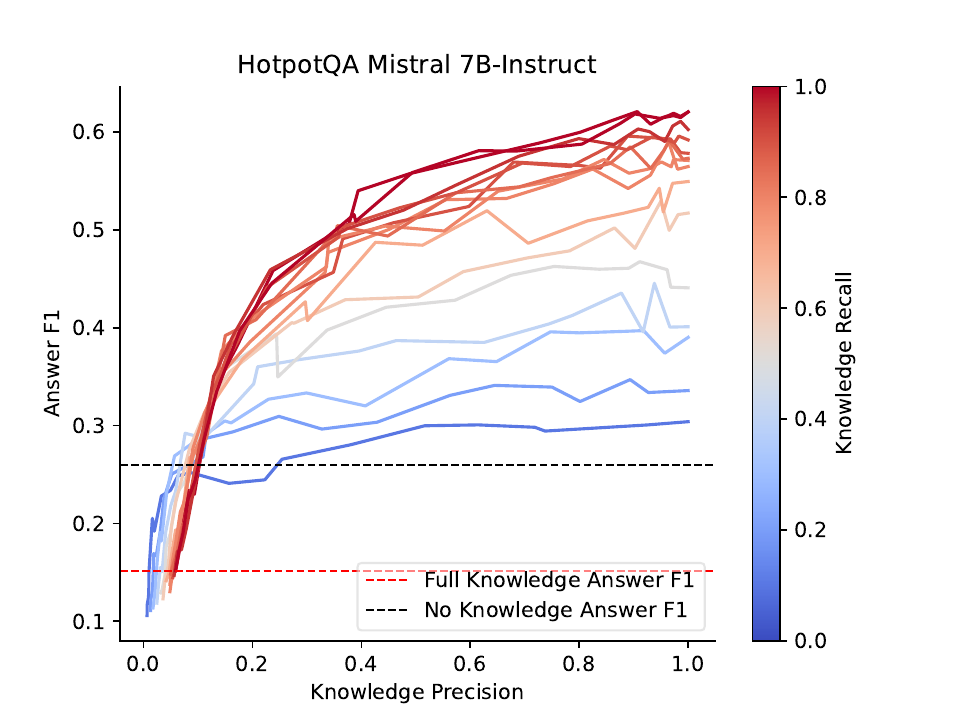}
    \label{fig:hotpot_qa_mistral_fixed_recall_precision_vs_answer_f1}
    \vspace{-1.5em} 
    \end{subfigure}

\vspace{-0.5em}
\caption{Color contours of answer F1 versus knowledge precision for GPT-4o-mini (top), LLaMA 3.1 8B (middle), and Mistral 7B-Instruct (bottom); the left column shows results on WoW, and the right shows HotpotQA. Each contour represents a different knowledge recall score; moving left to right visualizes improving the performance (precision) of the knowledge selector. Each plot point corresponds to a full experiment on the entire sampled dataset.}
\label{fig:fixed_recall_precision_vs_answer_f1}
 \vspace{-1.5em}
\end{figure*}

Overall we observe several consistent trends across meta-experiments, except for Mistral-7B-Instruct, which is a relatively weak-performing LLM. Note that each of the plots in Figures \ref{fig:selection_prec_recall_vs_answer_f1}-\ref{fig:fixed_recall_precision_vs_answer_f1} corresponds to a meta-experiment, and \textit{each point corresponds to a full experiment} on the entire test set. The standard error mean values in Tables \ref{tab:wow_benchmark} \& \ref{tab:hotpot_qa_benchmark} indicate that the answer F1 scores are robust across all experiments and the overall observed trends are significant.

\paragraph{RAG for LLM is beneficial.} As Tables \ref{tab:wow_benchmark} \& \ref{tab:hotpot_qa_benchmark} show, generators without any retrieved knowledge (``no knowledge") perform poorly on WoW and HotpotQA, even though they may have been pre-trained on the Wikipedia articles that WoW and HotpotQA are derived from. This finding indicates the LLMs are not over-fitted on the WoW and HotpotQA datasets, and applying RAG is beneficial. 

Interestingly, our HotpotQA results in Figure \ref{fig:selection_prec_recall_vs_answer_f1} show that distractor knowledge significantly harms performance; the cyan dots show that generators receiving mostly distractor knowledge underperform the ``no knowledge'' setting, which uses only the LLM's internal knowledge. This trend is not observed with WoW, supporting our observation that HotpotQA's distractor knowledge is truly irrelevant, while the ``distractor" knowledge in WoW may only be \textit{less} relevant than the single gold sentence, but not completely \textit{ir}relevant. This difference between datasets is further explored in Appendix \ref{sec:noisiness_of_wow}.

\paragraph{Full knowledge setting without knowledge selection is a strong baseline.}
Tables \ref{tab:wow_benchmark} \& \ref{tab:hotpot_qa_benchmark} show that the ``full knowledge" setting, which corresponds to knowledge retrieval with perfect recall, but no knowledge selection, is a very strong baseline. 

This finding mostly contrasts with those of prior knowledge selection works \cite{kimsequential, thulkeDSTC2021, li-etal-2022-enhancing-knowledge, sun-etal-2023-generative, zhang2023coarse, zheng2024ks, wang2024retrieve}. For example, in Table \ref{tab:hotpot_qa_benchmark}, we find that for GPT-4o-mini, a very strong generator model, the ``full knowledge" setting achieves 0.780 answer F1 on HotpotQA, only 0.048 lower than the ``gold knowledge" setting; similar observations are seen for LLaMA 3.1 8B on HotpotQA and for both models on WoW. Since the performance gap between the ``full knowledge'' and ``gold knowledge'' settings is the space for a knowledge selector to improve generation performance, we conclude that strong generators simply have less room for improvement via knowledge selection.

\paragraph{Knowledge precision \& recall together are good predictors of generation performance.} Figure \ref{fig:selection_prec_recall_vs_answer_f1} shows that generation performance varies smoothly with knowledge precision and recall, indicating that knowledge precision and recall together are major determinants of generation performance. Due to the pipeline nature of the retriever, selector, and generator components of RAG, knowledge \textit{recall} can only be improved by the \textit{retriever}. Meanwhile, the downstream \textit{selector} can only improve knowledge \textit{precision}, but it is likely to also \textit{reduce recall}. Therefore, we can visualize the effect of applying a knowledge selector: it moves the RAG performance down and to the right in Figure \ref{fig:selection_prec_recall_vs_answer_f1}; the better the knowledge selector, the more it will move \textit{right} (improving precision), and the less it will move \textit{down} (reducing recall).

\paragraph{Knowledge recall is the most crucial knowledge metric for strong generators.} 
We find that for strong generator models, the knowledge recall score is the best single knowledge metric for estimating generation performance; Figure \ref{fig:hotpotqa_selection_vs_answer_f1} shows a very strong correlation between knowledge recall and answer F1 for GPT-4o-mini on HotpotQA, and Figure \ref{fig:selection_recall_vs_answer_f1} in the Appendix shows the same for LLaMA 3.1 8B on HotpotQA and for both models on WoW. Moreover, knowledge recall is the most important factor in improving generation performance given a fixed generator, while the knowledge precision score is a secondary factor.

Figure \ref{fig:fixed_recall_precision_vs_answer_f1} further shows that increases in knowledge recall correspond to significant increases in answer F1 scores. Taking GPT-4o-mini on HotpotQA as an example, we see that moving from the left end of a color contour to the right (i.e. keeping knowledge recall fixed while improving precision) only slightly improves answer F1. In contrast, moving from one contour to another (i.e. varying knowledge recall) significantly impacts answer F1 for non-zero precision scores. Thus, to improve knowledge quality for RAG, improving the retriever's recall score is the top priority; improving precision via a knowledge selector has a limited contribution, especially for strong generators.

For weaker generators, like Mistral 7B-Instruct, the relationship between generation and knowledge F1 is stronger, and  correlation with recall is weaker (Figure \ref{fig:hotpotqa_selection_vs_answer_f1} right). We see much less separation between recall color contours and a much steeper increase with precision (Figure \ref{fig:fixed_recall_precision_vs_answer_f1} bottom).

\paragraph{Generator capability determines both overall performance and the usefulness of knowledge selection.}
All of our results show that the overall RAG performance depends on the generator model: the cross-model comparisons in Figures \ref{fig:selection_prec_recall_vs_answer_f1} \& \ref{fig:fixed_recall_precision_vs_answer_f1} show that for stronger generator models (as measured by rank in LLM leaderboards\footnote{\url{https://huggingface.co/spaces/lmsys/chatbot-arena-leaderboard}}), performance across all knowledge settings are higher (Tables \ref{tab:wow_benchmark} \& \ref{tab:hotpot_qa_benchmark}). While this result is unsurprising, we also find that the gap between the ``full knowledge'' and  ``gold knowledge'' settings becomes narrower, suggesting that stronger generators are more robust to noisy input knowledge and rely less on the knowledge selector. In contrast, when the generator is weaker, any reasonable knowledge selector becomes beneficial, likely because a weak generator cannot handle noisy input and requires a selector to filter out distractor knowledge; knowledge F1 best correlates with answer F1 for Mistral-7B-Instruct in Figure \ref{fig:hotpotqa_selection_vs_answer_f1}. 

\paragraph{Task and dataset are also key factors in RAG performance.} Figures \ref{fig:selection_prec_recall_vs_answer_f1} \& \ref{fig:fixed_recall_precision_vs_answer_f1} show that the same generator can show drastically different performance trends between WoW and HotpotQA. For example, while Mistral-7B-Instruct's performance degrades without a knowledge selector on HotpotQA, this is not the case for WoW. 

In addition, Figure \ref{fig:fixed_recall_precision_vs_answer_f1} shows that attempting to improve the knowledge selector on top of a weak retriever (i.e. increasing precision when recall is low) \textit{hurts} generation performance on WoW, as well as GPT-4o-mini on Hotpot QA. In this scenario, more total knowledge, regardless of noise, improves response generation. These counter-intuitive observations are likely due to the nature of the task and the annotation quality. The solution space in HotpotQA is small given a specific question, and HotpotQA has a cleaner separation between gold and distractor knowledge, whereas in WoW, there are more plausible responses given the same conversation history and knowledge, as well as ``distractor" knowledge that may actually be relevant to the gold response (see Appendix \ref{sec:noisiness_of_wow} for more analysis). 

\paragraph{Non-monotonic trends in improving the knowledge selector.} Interestingly, we observe that the boundary of where the knowledge selector improves generation performance (the border between the orange and white areas in Figure \ref{fig:selection_prec_recall_vs_answer_f1}) is convex for all three LLMs on WoW. This phenomenon can also be seen in Figure \ref{fig:fixed_recall_precision_vs_answer_f1}, where some color contours intersect with the ``full knowledge'' baseline (dashed red line) multiple times. In other words, we observe that generation performance is non-monotonic as knowledge precision increases with a fixed knowledge recall. 
The fact that this phenomenon is only observed on WoW may be due to the relatively noisy gold knowledge annotations in WoW; we can produce similar behavior by artificially injecting noise into HotpotQA's gold knowledge annotations (Appendix \ref{sec:noisy_hotpotqa}).

\paragraph{Constraining the knowledge size does not change generation accuracy.} One important motivation for using a knowledge selector is to reduce the total input length to the generator model. However, while computational costs can be reduced by only using the top-$k$ knowledge, we find that the overall relationship between knowledge precision-recall and generation F1 observed in our simulations remains unchanged, regardless of the value of $k$ (Appendix \ref{sec:length_limit}). 

\section{Conclusion and Discussion}

In this study, we systematically examined the behavior of RAG generation in relation to the performance of knowledge retrieval and selection. 

\subsection{Overall Observations}
Summing up Section \ref{sec:RagSimulation_results}, we find an interaction effect between generator capability and ambiguity of the task/dataset on RAG generation performance. We find two types of generator behavior:

\textbf{A strong generator model} can achieve good performance without a knowledge selector because it is robust to noise, and therefore performs better as more gold knowledge is retrieved, despite the presence of distractor knowledge. Thus, knowledge recall correlates well with generation F1, and there is less room for improvement through the addition of a knowledge selector.

\textbf{A weak generator model} cannot handle distractor knowledge and requires a selector to refine the noisy input knowledge, and thus knowledge F1 correlates well with generation F1. Since nowadays most popular generator models are strong LLMs, knowledge selectors have a limited benefit. We hypothesize that prior work that found performance improvements from dedicated knowledge selectors saw those benefits because their generator models were weak; most prior work used BART \cite{lewis-etal-2020-bart} as the generator.

Finally, we note that the strength of a generator model is relative; even a SOTA generator can fall into the ``weak" category, given a sufficiently noisy and challenging task/dataset. We hope the visualizations shown in our figures can help guide future practitioners improve their RAG systems in real-world applications. 

\subsection{Recommendations to Practitioners}
Based on our observations, we make the following recommendations for future practitioners considering using knowledge selectors to improve performance in a real-world RAG scenario:

Benchmark the generation performance without external knowledge, with all candidate knowledge, and if possible, with gold knowledge only. The ``no knowledge'' setting measures the generator's base performance, ``full knowledge'' serves as a strong baseline corresponding to knowledge retrieval with perfect recall, and ``gold knowledge'' gives the upper bound of RAG performance with perfect knowledge selection. The gap between ``full knowledge" and ``gold knowledge" is the potential performance gain brought by adding a knowledge selector.


To improve RAG performance, increasing the knowledge recall score is the most effective strategy. 
Thus, for modern LLM generators, improving the knowledge retriever's recall is key. In practice, the retriever or selector may encounter false-negative gold knowledge during training, as is the case with WoW, which makes prioritizing recall even more important. Moreover, because modern LLM generators have long maximum input windows (e.g. 128k for GPT-4o \& -mini), having too much knowledge is less likely to be a problem. If the number of knowledge sentences must be constrained to reduce computational cost, the top-$k$ knowledge sentences should maintain the same level of knowledge recall and precision as the non-length-constrained settings.

Only when the knowledge recall score is high can a knowledge selector, which increases knowledge precision, be potentially helpful\footnote{In a length-constrained scenario, if we consider $k$  randomly sampled knowledge sentences as a baseline, rather than the ``full knowledge'' setting, then the knowledge selector is more likely to be helpful due to weaker baseline performance.}. Moreover, if the recall is too low, the downstream generation performance may not see any benefits from a knowledge selector with middling performance --- it may even be harmed; only when the knowledge selector's precision is very high will the overall performance improve.


\section*{Limitations}
\paragraph{Computational resources for simulations.} Due to the high cost of larger-scale simulations in each meta-experiment that consists of hundreds of full experiments using API-based LLMs, we use only a subset of the WoW and HotpotQA for experiments. As a result, our data subset may contain some minor noise during knowledge sampling and LLM generation, and the contours in Figure \ref{fig:fixed_recall_precision_vs_answer_f1} are not smooth. However, such noise is not likely to affect our conclusions. For a similar reason, we only chose three LLMs for our experiments. As a result, we may have missed out on more subtle phenomena that can only be seen from results on a larger number of generators.

\paragraph{Datasets for in-depth analysis.} In addition to the cost issue, there are very few RAG datasets with human-annotated knowledge \cite{friel2024ragbench}, which further limits our simulation experiment settings. Furthermore, even though we regard WoW as a relatively noisy dataset in this work compared to HotpotQA, to the best of our knowledge, WoW is one of the most cleanly annotated datasets among all datasets. We cannot verify our hypothesis in Section \ref{sec:RagSimulation_results} that WoW has a larger solution space than HotpotQA without re-annotating the dataset because each example in WoW only contains one gold response. Our experiments in Appendix \ref{sec:noisiness_of_wow} compare the noisiness of gold knowledge annotations in WoW with those in HotpotQA.

\paragraph{Uniform Sampling Probably for Simulation.} In Section \ref{sec:simulation_approach}, we draw gold and distractor knowledge from a uniform distribution $p_{gold}$ and $p_{noise}$ for simplicity since we do not have a prior assumption of the knowledge selector's preference. However, a real knowledge selector may be more likely to select one knowledge sentence than another. Nonetheless, we successfully identified the knowledge precision and recall scores together as good predictors of the generation performance as we analyzed in Section \ref{sec:RagSimulation_results}.

\section*{Ethics Statement}
Since all datasets and LLMs used in this work are publicly available or accessible, and no data collection introducing sensitive information is performed, the ethical considerations of this study are minimal.

\clearpage
\bibliography{custom}
\clearpage
\appendix

\section{Appendix}
\label{sec:appendix}

\subsection{Implementation Details} \label{sec:implementation_detail}
\subsubsection{Knowledge Sampling} \label{sec:sampling_detail}
Since knowledge precision and recall are the most common metrics for knowledge retrieval and selection performance, we use these metrics as the basis for our analysis. We find that the sampling rate $p_{gold}$ linearly correlates with knowledge recall scores, while $p_{noise}$ exponentially correlates with knowledge precision. Thus, to simulate retrieving and selecting a knowledge set $K'$ with a specific knowledge precision and recall score, we use grid search in the linear space of $p_{gold}$ and both the linear and exponential spaces of $p_{noise}$ to ensure most grids in the knowledge precision-recall space are covered by our experiments. 

We maintain the original order of the knowledge sentences from the documents provided in the datasets and do not observe a strong influence from the position of the gold knowledge sentences.

\subsubsection{Models}
We do not perform any fine-tuning or hyperparameter tuning. We set the LLMs' temperature to 0 and ensure that our zero-shot generation prompts (Appendix \ref{sec:prompts}) are shorter than their maximum input lengths. We use the API services of OpenAI gpt-4o-mini-2024-07-18\footnote{\url{https://platform.openai.com/docs/models/gpt-4o-mini}}, Together AI\footnote{\url{https://docs.together.ai/docs/chat-models}} meta-llama/Meta-Llama-3.1-8B-Instruct-Turbo, and mistralai/Mistral-7B-Instruct-v0.1. We report results on the first 500 examples in the HotpotQA training set and the first 100 conversations (452 wizard utterances) from the ``test seen'' set of WoW. Our set of experiments cost about 50 USD from OpenAI and about 50 USD from Together.ai.

\subsection{Prompts} \label{sec:prompts}
Tables \ref{tab:WoW_prompt} \& \ref{tab:hotpotQA_prompt} show the prompts we use for response generation on WoW and HotpotQA, respectively. 
While we experimented with Chain-of-Thought \cite{wei2022chain} for these prompts, they did not outperform zero-shot prompting for LLaMA 3.1 8B and Mistral-7B-Instruct, so we use zero-shot prompting throughout our experiments to keep the settings as simple as possible.

\subsection{Additional Plots}
\begin{figure*}[t]
     \centering
        \begin{subfigure}[b]{0.48\textwidth}
        \centering
          \includegraphics[width=\textwidth]{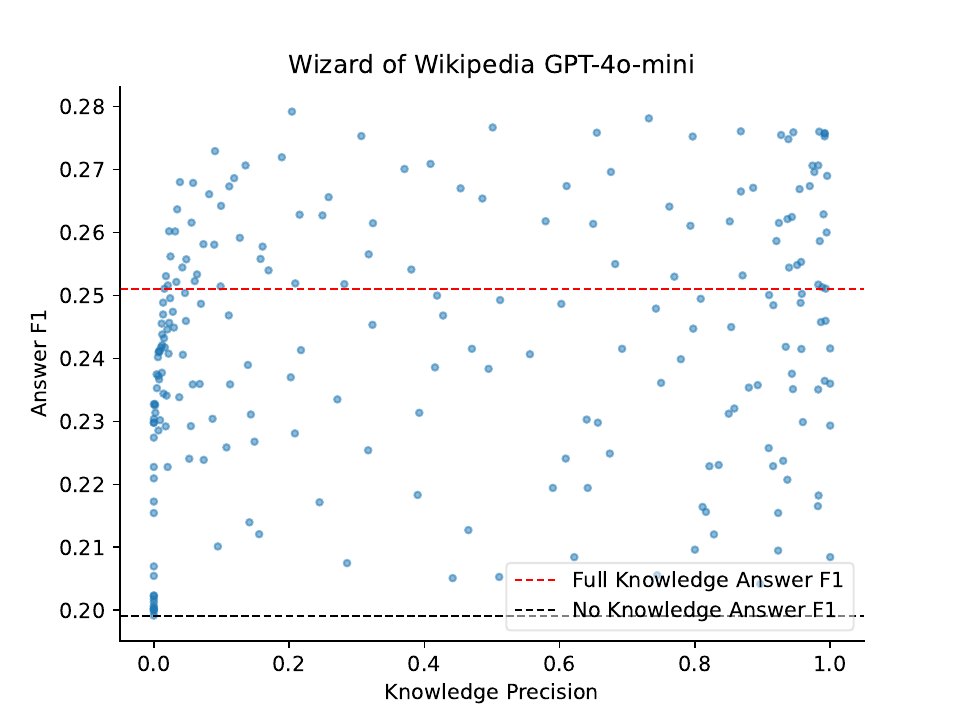}
          \label{fig:wow_gpt_selection_prec_vs_answer_f1}
          \vspace{-1.5em}
        \end{subfigure}
        \quad
        \begin{subfigure}[b]{0.48\textwidth}
        \centering
          \includegraphics[width=\textwidth]{GPT-4o-mini_HotpotQA_figures/selection_prec_vs_answer_f1.pdf}
          \label{fig:hotpotqa_gpt_selection_prec_vs_answer_f1}
          \vspace{-1.5em}
        \end{subfigure}
     \quad
        \begin{subfigure}[b]{0.48\textwidth}
        \centering
          \includegraphics[width=\textwidth]{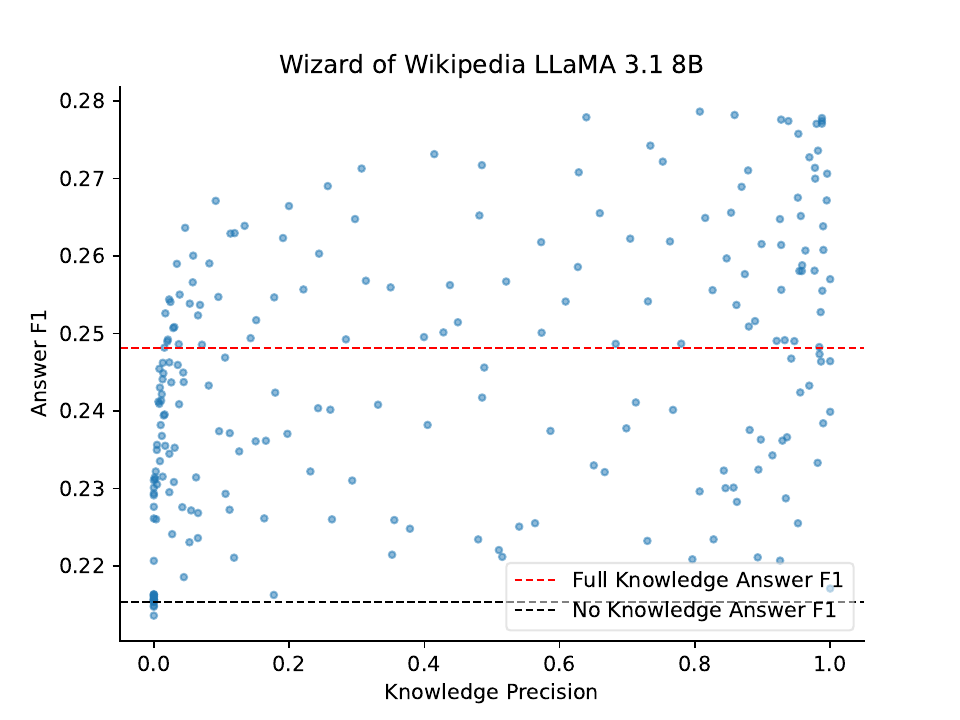}
          \label{fig:wow_llama_selection_prec_vs_answer_f1}
          \vspace{-1.5em}
        \end{subfigure}
     \quad
         \begin{subfigure}[b]{0.48\textwidth}
        \centering
          \includegraphics[width=\textwidth]{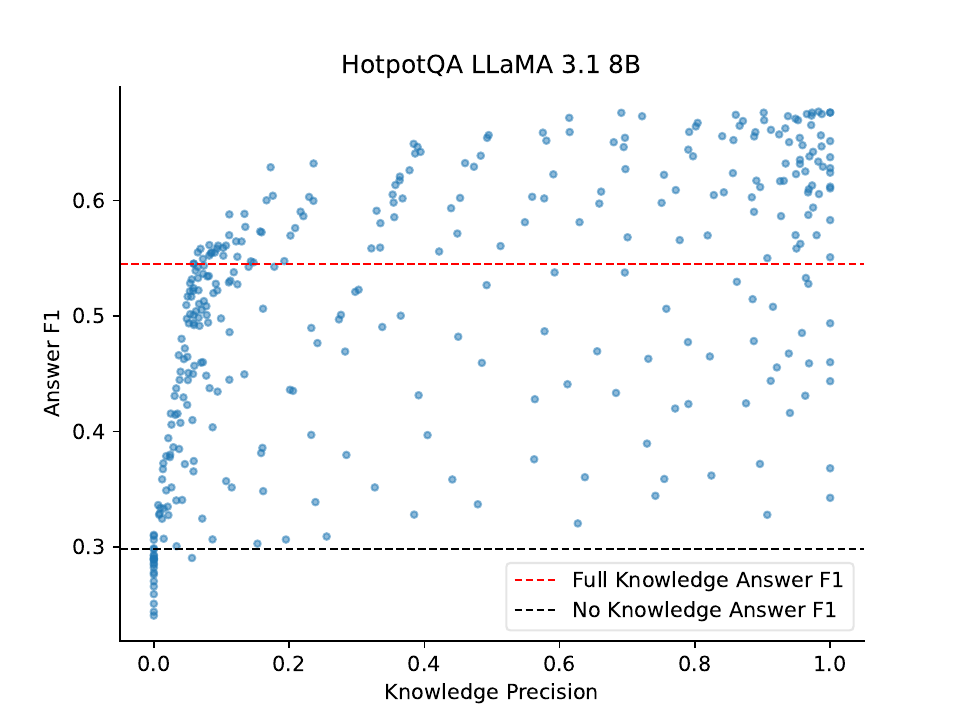}
          \label{fig:hotpotqa_llama_selection_prec_vs_answer_f1}
          \vspace{-1.5em}
        \end{subfigure}
     \quad
        \begin{subfigure}[b]{0.48\textwidth}
        \centering
          \includegraphics[width=\textwidth]{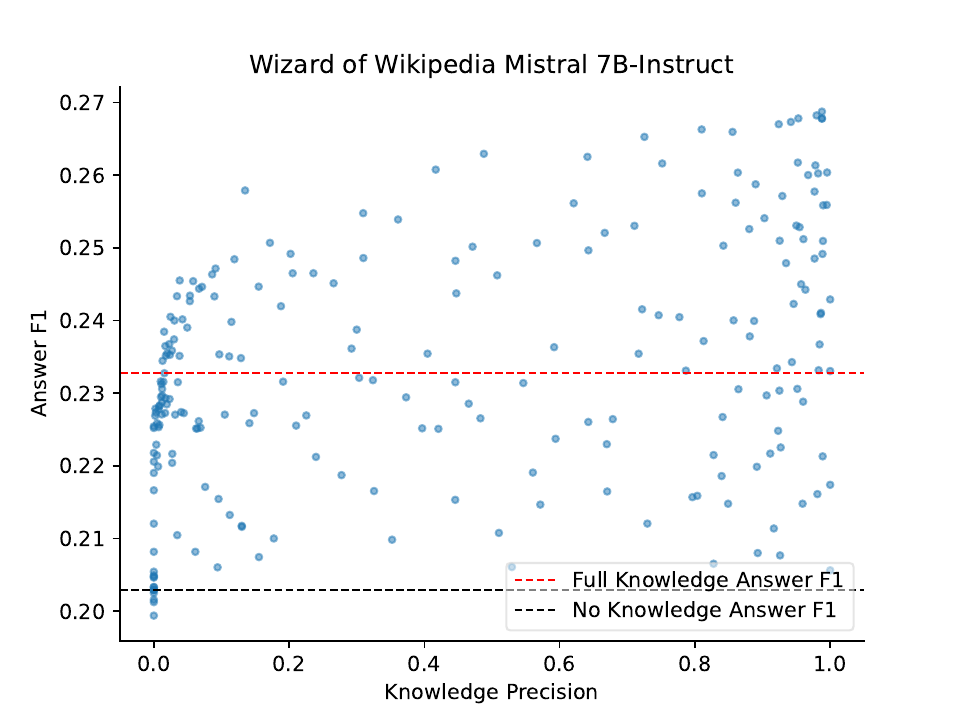}
          \label{fig:wow_mistral_selection_prec_vs_answer_f1}
          \vspace{-1.5em}
        \end{subfigure}
     \quad
     \begin{subfigure}[b]{0.48\textwidth}
        \centering
          \includegraphics[width=\textwidth]{Mistral_7B_HotpotQA_figures/selection_prec_vs_answer_f1.pdf}
          \label{fig:hotpotqa_mistral_selection_prec_vs_answer_f1}
          \vspace{-1.5em}
        \end{subfigure}
     \quad
     \vspace{-0.5em}
        \caption{Scatter plot of answer F1 versus the knowledge precision for GPT-4o-mini (top), LLaMA 3.1 8B (middle), and Mistral 7B-Instruct (bottom); the left column shows results on WoW, and the right shows HotpotQA. \textit{Each data point corresponds to a full experiment on the entire sampled dataset.}}
        \label{fig:selection_precision_vs_answer_f1}
\end{figure*}

\begin{figure*}[t]
     \centering
        \begin{subfigure}[b]{0.48\textwidth}
        \centering
          \includegraphics[width=\textwidth]{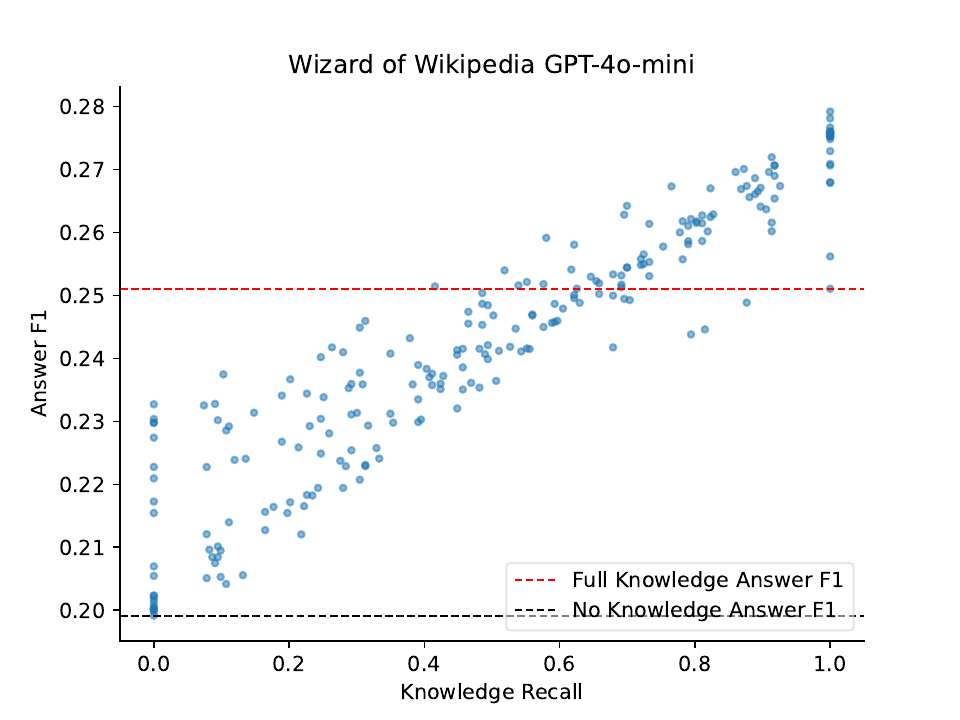}
          \label{fig:wow_gpt_selection_recall_vs_answer_f1}
          \vspace{-1.5em}
        \end{subfigure}
        \quad
        \begin{subfigure}[b]{0.48\textwidth}
        \centering
          \includegraphics[width=\textwidth]{GPT-4o-mini_HotpotQA_figures/selection_recall_vs_answer_f1.pdf}
          \label{fig:hotpotqa_gpt_selection_recall_vs_answer_f1}
          \vspace{-1.5em}
        \end{subfigure}
     \quad
        \begin{subfigure}[b]{0.48\textwidth}
        \centering
          \includegraphics[width=\textwidth]{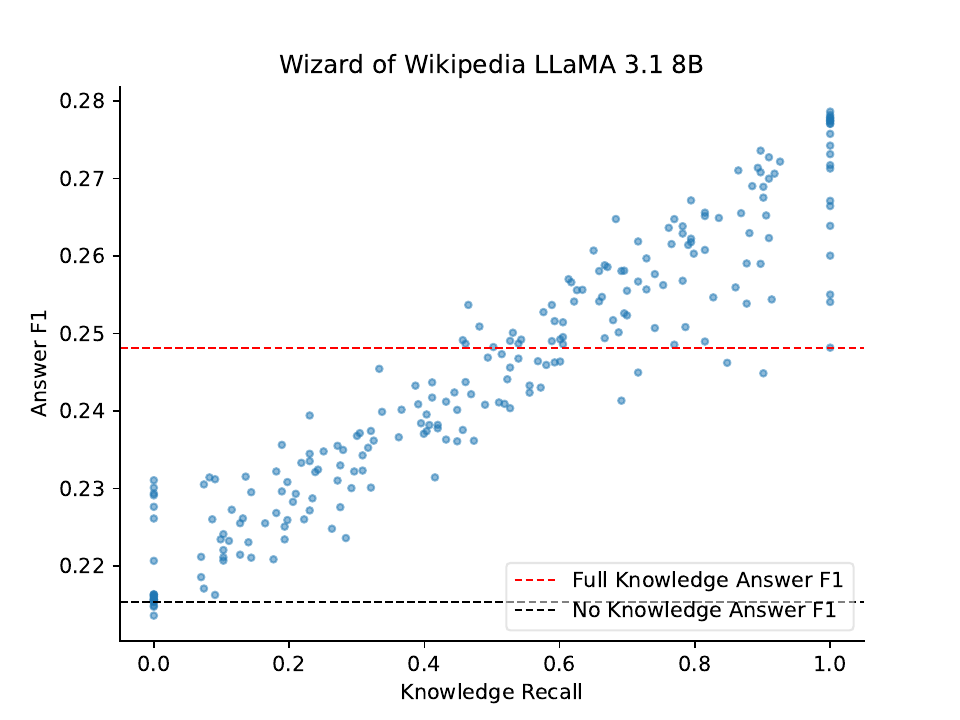}
          \label{fig:wow_llama_selection_recall_vs_answer_f1}
          \vspace{-1.5em}
        \end{subfigure}
     \quad
         \begin{subfigure}[b]{0.48\textwidth}
        \centering
          \includegraphics[width=\textwidth]{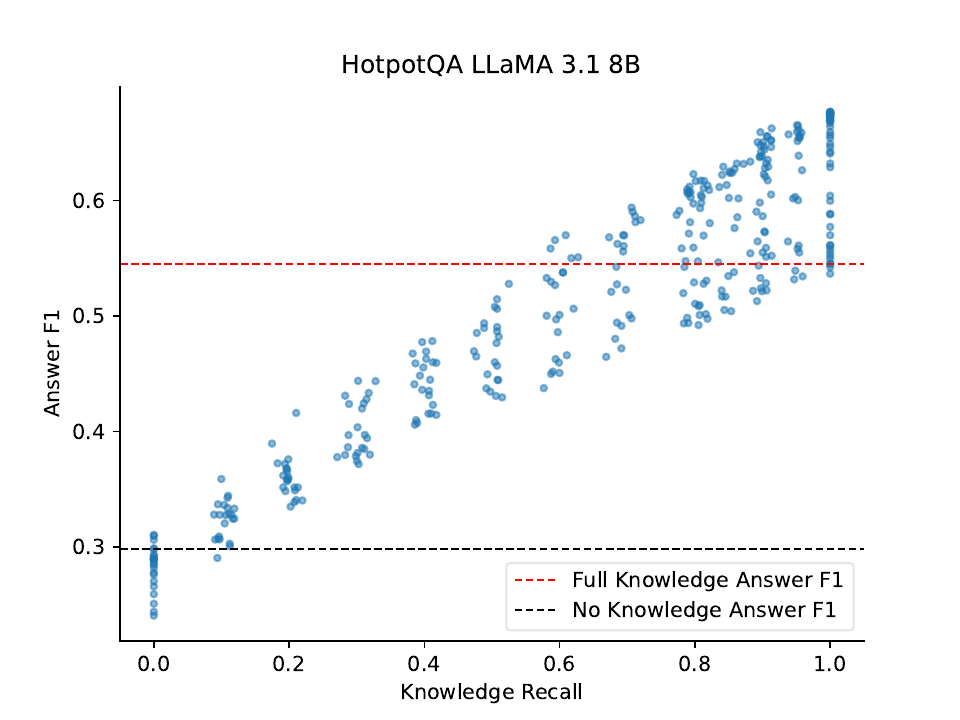}
          \label{fig:hotpotqa_llama_selection_recall_vs_answer_f1}
          \vspace{-1.5em}
        \end{subfigure}
     \quad
        \begin{subfigure}[b]{0.48\textwidth}
        \centering
          \includegraphics[width=\textwidth]{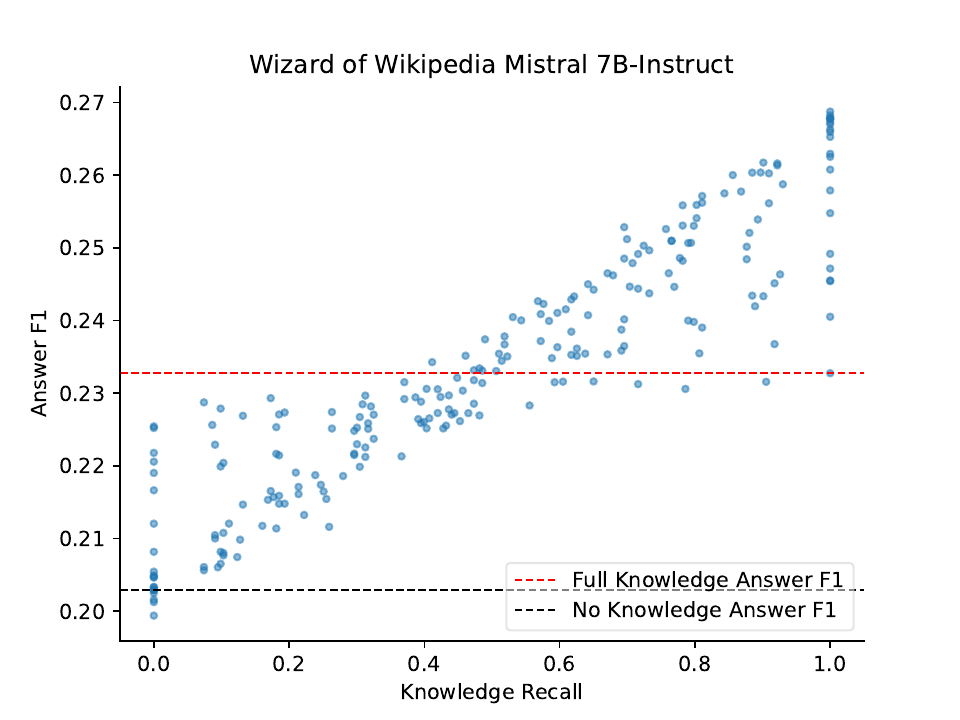}
          \label{fig:wow_mistral_selection_recall_vs_answer_f1}
          \vspace{-1.5em}
        \end{subfigure}
     \quad
     \begin{subfigure}[b]{0.48\textwidth}
        \centering
          \includegraphics[width=\textwidth]{Mistral_7B_HotpotQA_figures/selection_recall_vs_answer_f1.pdf}
          \label{fig:hotpotqa_mistral_selection_recall_vs_answer_f1}
          \vspace{-1.5em}
        \end{subfigure}
     \quad
     \vspace{-0.5em}
        \caption{Scatter plot of answer F1 versus the knowledge recall for GPT-4o-mini (top), LLaMA 3.1 8B (middle), and Mistral 7B-Instruct (bottom); the left column shows results on WoW, and the right shows HotpotQA. \textit{Each data point corresponds to a full experiment on the entire sampled dataset.}}
        \label{fig:selection_recall_vs_answer_f1}
\end{figure*}

\begin{figure*}[t]
     \centering
        \begin{subfigure}[b]{0.48\textwidth}
        \centering
          \includegraphics[width=\textwidth]{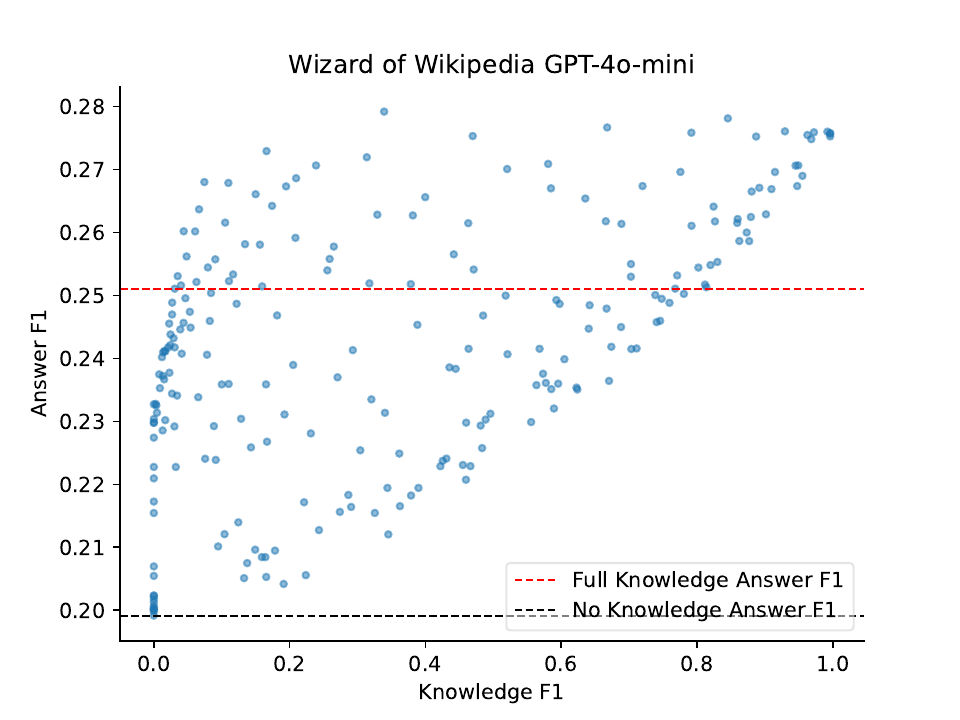}
          \label{fig:wow_gpt_selection_f1_vs_answer_f1}
          \vspace{-1.5em}
        \end{subfigure}
        \quad
        \begin{subfigure}[b]{0.48\textwidth}
        \centering
          \includegraphics[width=\textwidth]{GPT-4o-mini_HotpotQA_figures/selection_f1_vs_answer_f1.pdf}
          \label{fig:hotpotqa_gpt_selection_f1_vs_answer_f1}
          \vspace{-1.5em}
        \end{subfigure}
     \quad
        \begin{subfigure}[b]{0.48\textwidth}
        \centering
          \includegraphics[width=\textwidth]{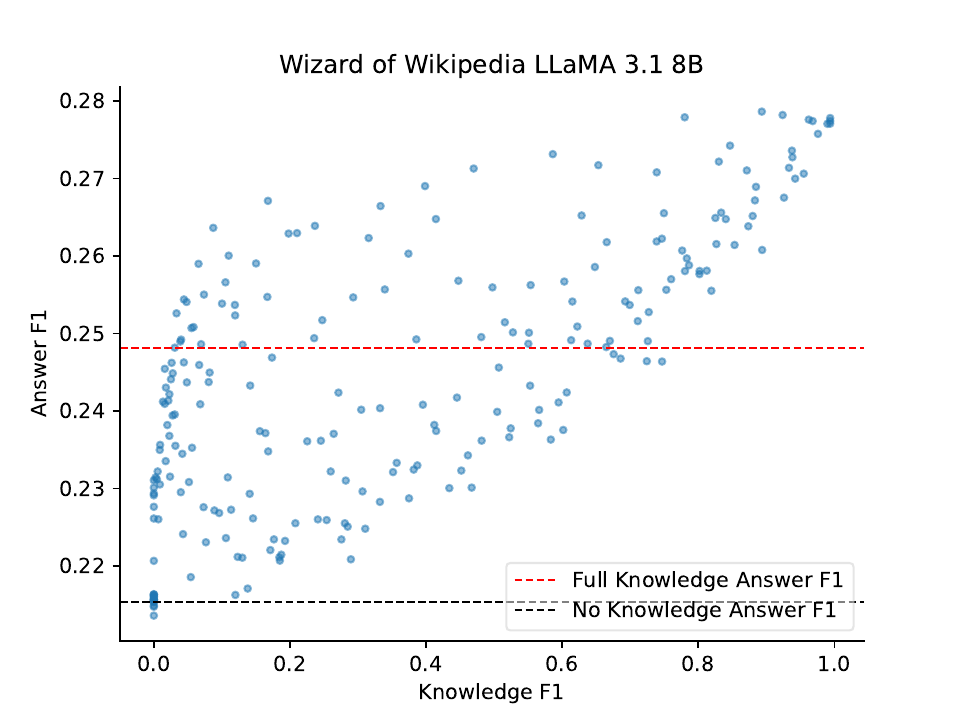}
          \label{fig:wow_llama_selection_f1_vs_answer_f1}
          \vspace{-1.5em}
        \end{subfigure}
     \quad
         \begin{subfigure}[b]{0.48\textwidth}
        \centering
          \includegraphics[width=\textwidth]{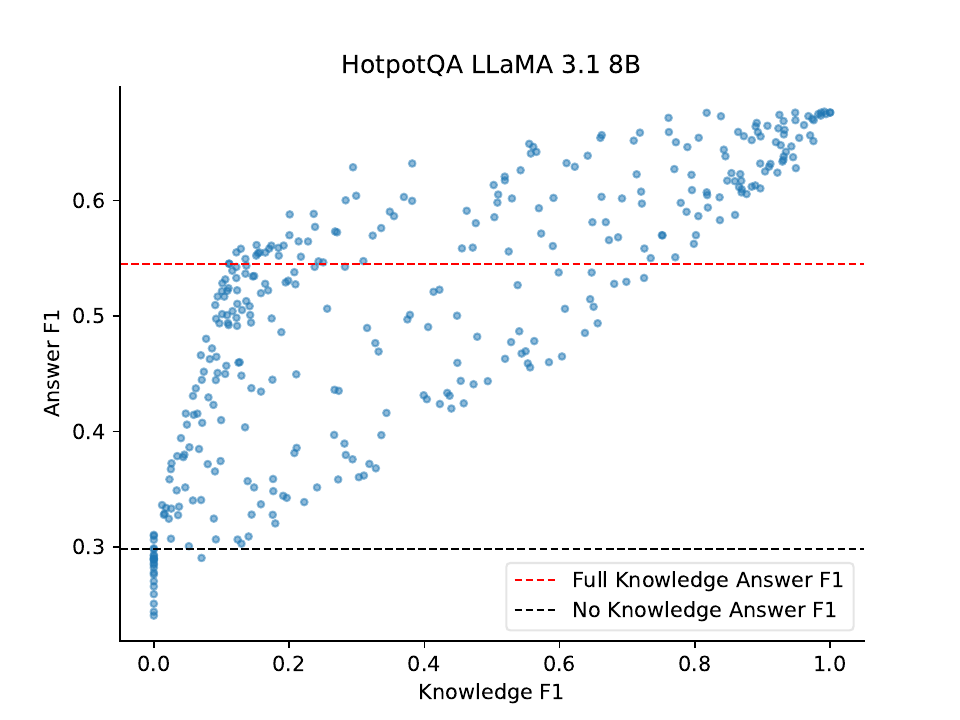}
          \label{fig:hotpotqa_llama_selection_f1_vs_answer_f1}
          \vspace{-1.5em}
        \end{subfigure}
     \quad
        \begin{subfigure}[b]{0.48\textwidth}
        \centering
          \includegraphics[width=\textwidth]{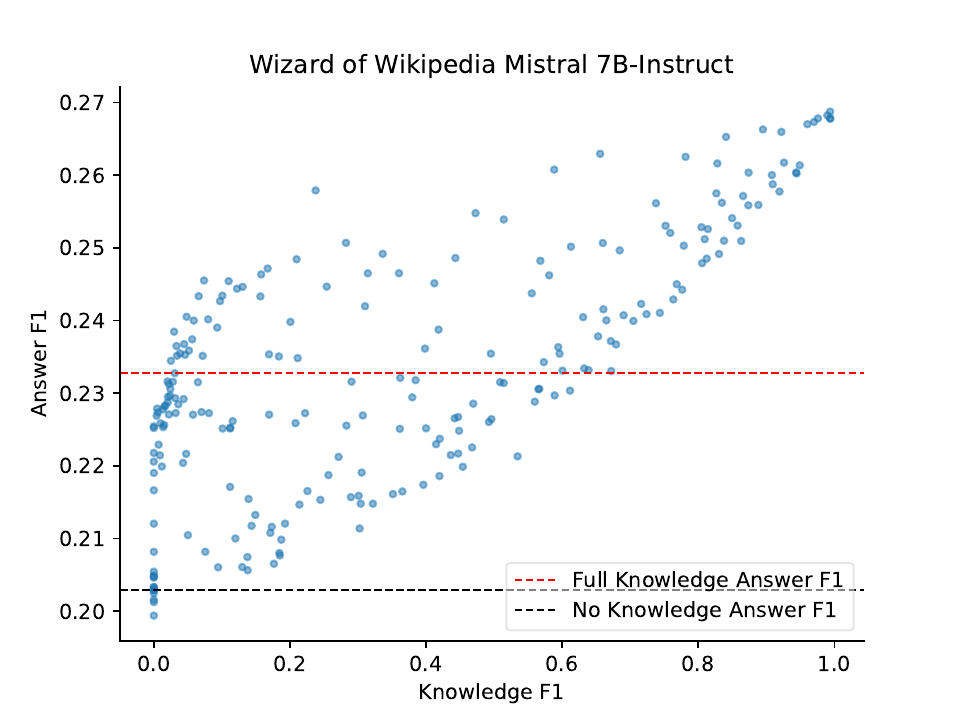}
          \label{fig:wow_mistral_selection_f1_vs_answer_f1}
          \vspace{-1.5em}
        \end{subfigure}
     \quad
     \begin{subfigure}[b]{0.48\textwidth}
        \centering
          \includegraphics[width=\textwidth]{Mistral_7B_HotpotQA_figures/selection_f1_vs_answer_f1.pdf}
          \label{fig:hotpotqa_mistral_selection_f1_vs_answer_f1}
          \vspace{-1.5em}
        \end{subfigure}
     \quad
     \vspace{-0.5em}
        \caption{Scatter plot of answer F1 versus the knowledge F1 for GPT-4o-mini (top), LLaMA 3.1 8B (middle), and Mistral 7B-Instruct (bottom); the left column shows results on WoW, and the right shows HotpotQA. \textit{Each data point corresponds to a full experiment on the entire sampled dataset.}}
        \label{fig:selection_f1_vs_answer_f1}
\end{figure*}

Figures \ref{fig:selection_precision_vs_answer_f1}, \ref{fig:selection_recall_vs_answer_f1}, and \ref{fig:selection_f1_vs_answer_f1} extend Figure \ref{fig:hotpotqa_selection_vs_answer_f1}, with knowledge precision, knowledge recall, and knowledge F1 plotted against response/answer F1 for WoW and HotpotQA, respectively, for all three LLM generators. Figure \ref{fig:selection_recall_vs_answer_f1} shows that knowledge recall correlates strongly with response/answer F1 for both GPT-4o-mini and LLaMA 3.1 8B on both WoW and HotpotQA, as well as for Mistral 7B-Instruct on WoW, while Figure \ref{fig:selection_f1_vs_answer_f1} shows that knowledge F1 has the stronger correlation for Mistral 7B-Instruct on HotpotQA.

\subsection{Noisiness of Wizard of Wikipedia vs. HotpotQA} \label{sec:noisiness_of_wow}
\begin{figure}[t]
 \centering
     \includegraphics[width=0.48\textwidth]{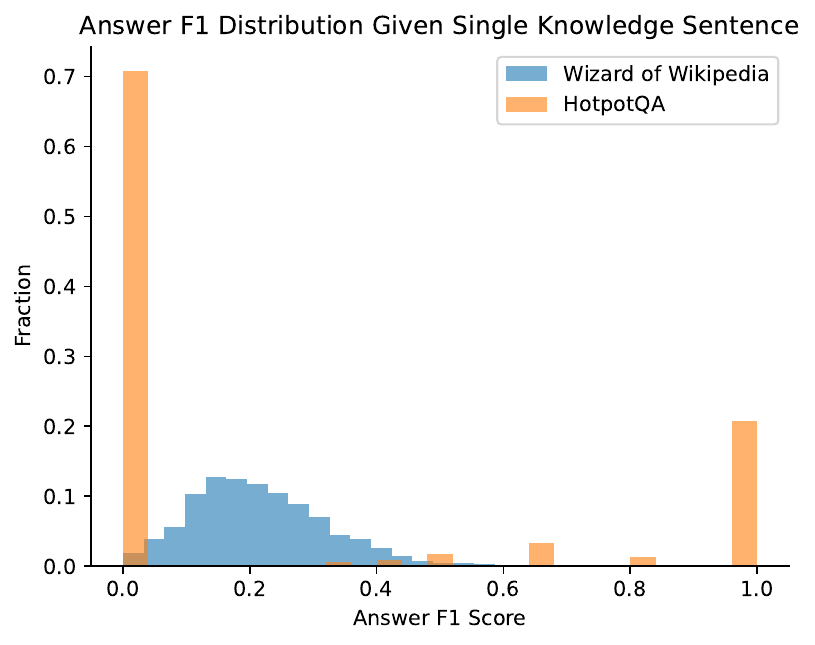}
    \caption{Histogram of answer F1 distributions of Wizard of Wikipedia and HotpotQA by feeding each \textit{individual} candidate sentence to the GPT-4o-mini generator.} 
    \label{fig:answer_f1_distribution}
\end{figure}

To intuitively compare the noisiness of WoW vs. HotpotQA, for each example, we feed each \textit{individual} candidate knowledge sentence, regardless of gold or distractor status, to the GPT-4o-mini generator model and measure the answer F1 score to measure how each individual knowledge sentence affects the generation performance. As Figure \ref{fig:answer_f1_distribution} shows, WoW knowledge sentences result in a wide, continuous distribution of response F1 scores, indicating that even the ``distractor" knowledge positively contributes to the gold response to some extent. In contrast, we find that 70\% of HotpotQA sentences are true distractors that cause the generator to produce incorrect answers with 0 F1 score, while 20\% of sentences result in correct answers. These results indicate that WoW is a much noisier dataset than HotpotQA.

\subsection{Noisy HotpotQA} \label{sec:noisy_hotpotqa}
As we discuss in Section \ref{sec:datasets}, the WoW gold knowledge annotations are noisy in that only one sentence can be seelcted as gold knowledge, even if the other retrieved sentences are still relevant. To test the impact of such knowledge annotation noise on our RAG simulation results, we deliberately inject noise into HotpotQA by removing one sentence from the gold knowledge set for each example to create a \textit{noisy HotpotQA} dataset. As a result, the average number of gold knowledge sentences per example drops from 2.4 to 1.4, and a small portion of the ``distractor" knowledge in the noisy dataset is actually gold knowledge from the original HotpotQA (i.e. false-negative knowledge).

\begin{figure}[t]
\centering
  \includegraphics[width=0.48\textwidth]{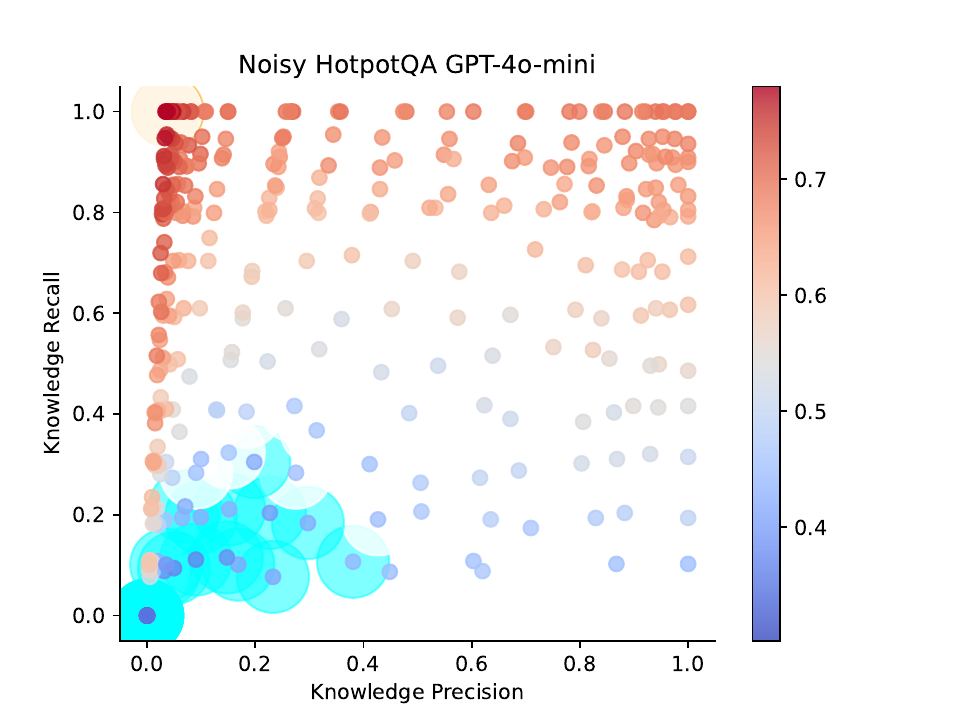}
  \caption{Scatter plot of answer F1 versus the knowledge precision for GPT-4o-mini on \textbf{noisy} HotpotQA.} 
  \label{fig:noisy_hotpot_qa_gpt_selection_prec_recall_by_area}
\end{figure}

\begin{figure}[t]
 \centering
     \includegraphics[width=0.48\textwidth]{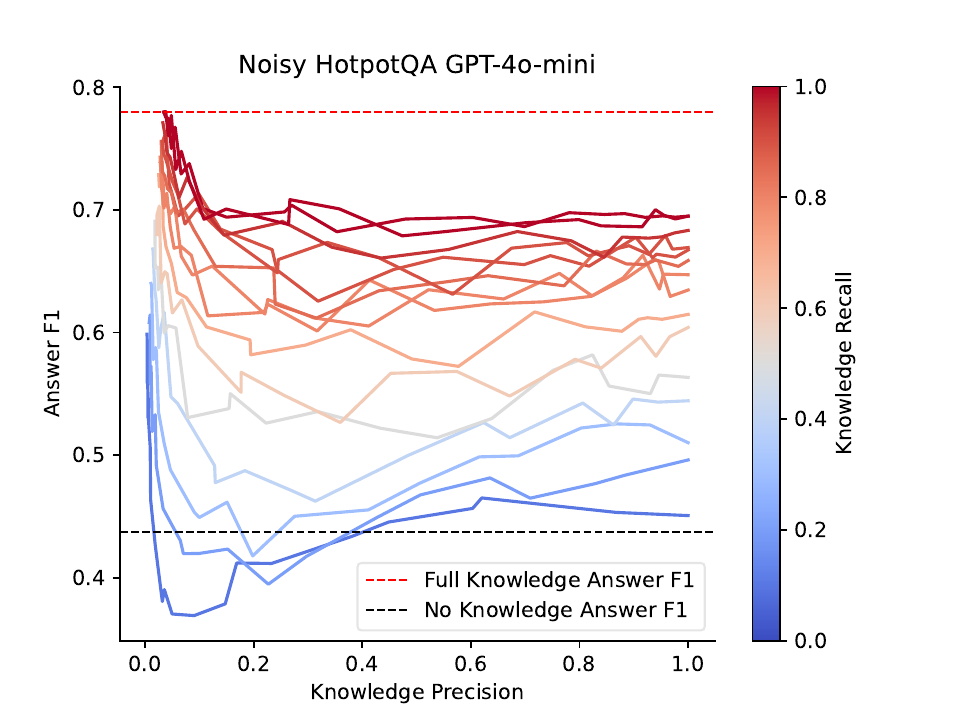}
    \caption{Color contours of answer F1 versus the knowledge precision for GPT-4o-mini on \textbf{noisy} HotpotQA. Each contour represents a different knowledge recall score; moving left to right visualizes improving the performance (precision) of the knowledge selector.} 
    \label{fig:noisy_hotpot_qa_gpt_fixed_recall_precision_vs_answer_f1}
\end{figure}

As Figures \ref{fig:noisy_hotpot_qa_gpt_selection_prec_recall_by_area} \& \ref{fig:noisy_hotpot_qa_gpt_fixed_recall_precision_vs_answer_f1} show, the injected noise drastically impairs the knowledge selector's efficacy: even given a ``perfect'' knowledge selector with a knowledge precision score of 100\%, the answer F1 still underperforms the ``full knowledge'' setting. This is because our knowledge precision and recall scores are computed based on the labeled gold knowledge sentences, so if the labels are noisy, the knowledge metrics are corrupted, while the answer F1 is intact. Note that Figure \ref{fig:noisy_hotpot_qa_gpt_fixed_recall_precision_vs_answer_f1} presents a non-monotonic trend in improving knowledge selector, just as we observed in Section \ref{sec:RagSimulation_results} for WoW, verifying that noisy knowledge annotations cause this non-monotonic behavior.

We argue that the simulation results on WoW and noisy HotpotQA are more reflective of real-world RAG applications than the original (clean) HotpotQA. In a real application, there is no way to know the test-time knowledge recall and precision, and it is likely to be slightly different than the training-time knowledge recall and precision. 

\subsection{Length-Constrained Knowledge Selection and Answer Generation} \label{sec:length_limit}
To study the effect of length-constrained knowledge inputs on answer generation performance, we limit the number of knowledge sentences to $k$ by randomly sub-sampling $k$ sentences when more than $k$ sentences are selected by the simulated knowledge selector\footnote{This random sampling does not break the experiment because our knowledge retriever and selector are simulated via random sampling, which is not the case for actual model-based retrievers and selectors.}. We choose $k=3$ for both WoW and HotpotQA because their average number of candidate knowledge sentences \textit{before random sampling} are 7.1 and 9.5 respectively.

\begin{figure*}
     \centering
        \begin{subfigure}[b]{0.48\textwidth}
        \centering
          \includegraphics[width=\textwidth]{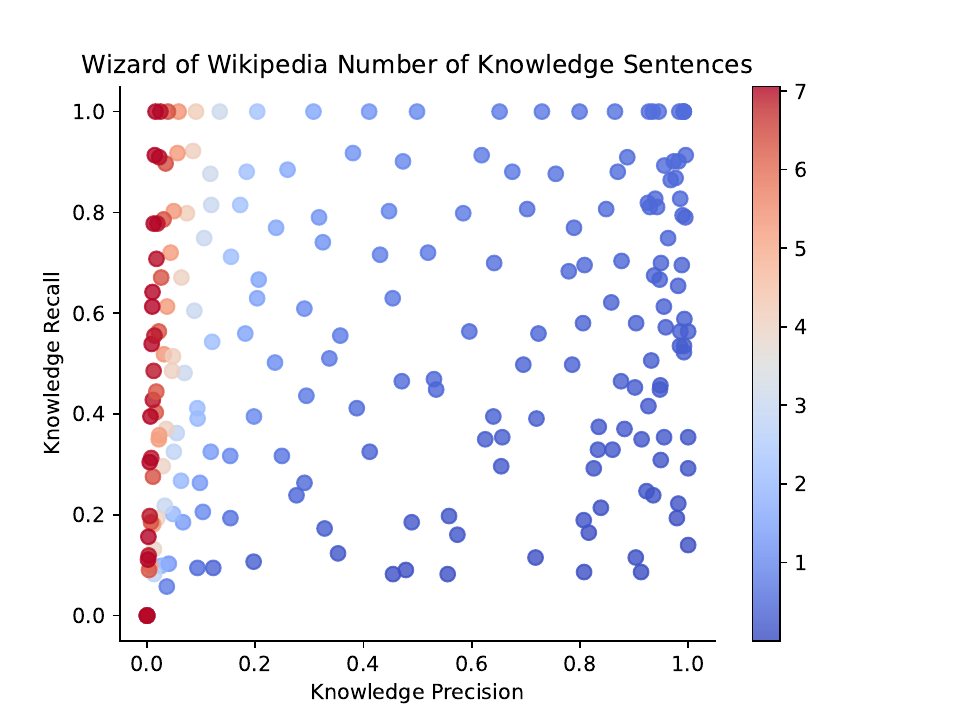}
          \label{fig:wow_gpt_selection_prec_recall_vs_k}
          \vspace{-1.5em}
        \end{subfigure}
        \quad
      \begin{subfigure}[b]{0.48\textwidth}
        \centering
          \includegraphics[width=\textwidth]{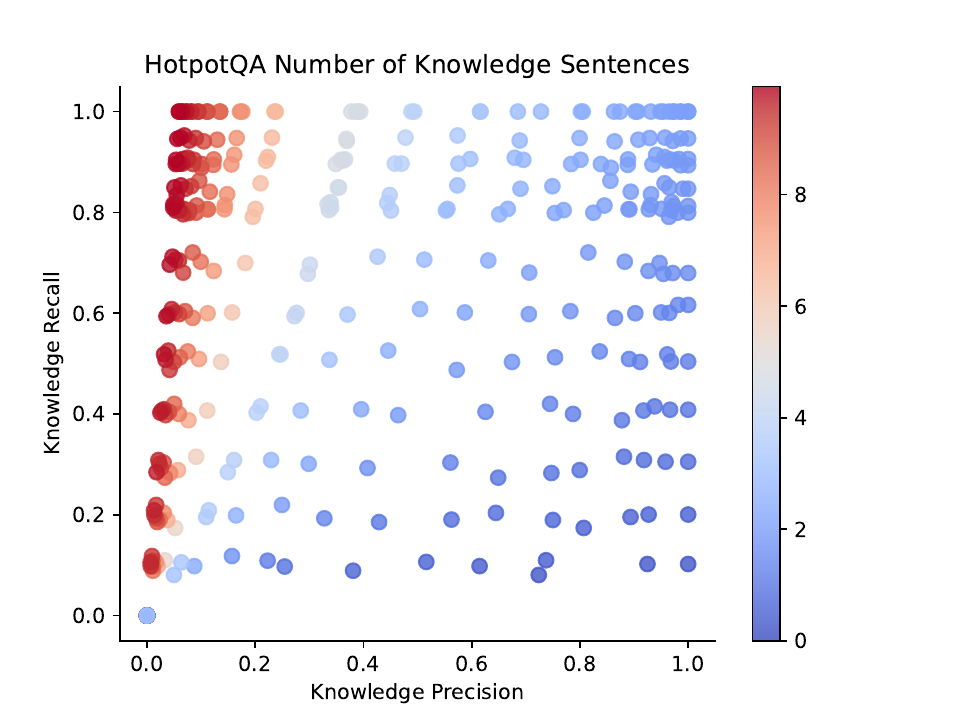}
          \label{fig:hotpot_qa_gpt_selection_prec_recall_vs_k}
          \vspace{-1.5em}
        \end{subfigure}
     \quad
       
     \vspace{-0.5em}
        \caption{Scatter plot of the number of selected knowledge sentences versus the knowledge precision on Wizard of Wikipedia (left) and HotpotQA (right). \textit{Each data point corresponds to a full experiment on the entire sampled dataset.}}
        \label{fig:knowledge_selection_precision_recall_vs_k}
\end{figure*}

\begin{figure*}
     \centering
        \begin{subfigure}[b]{0.48\textwidth}
        \centering
          \includegraphics[width=\textwidth]{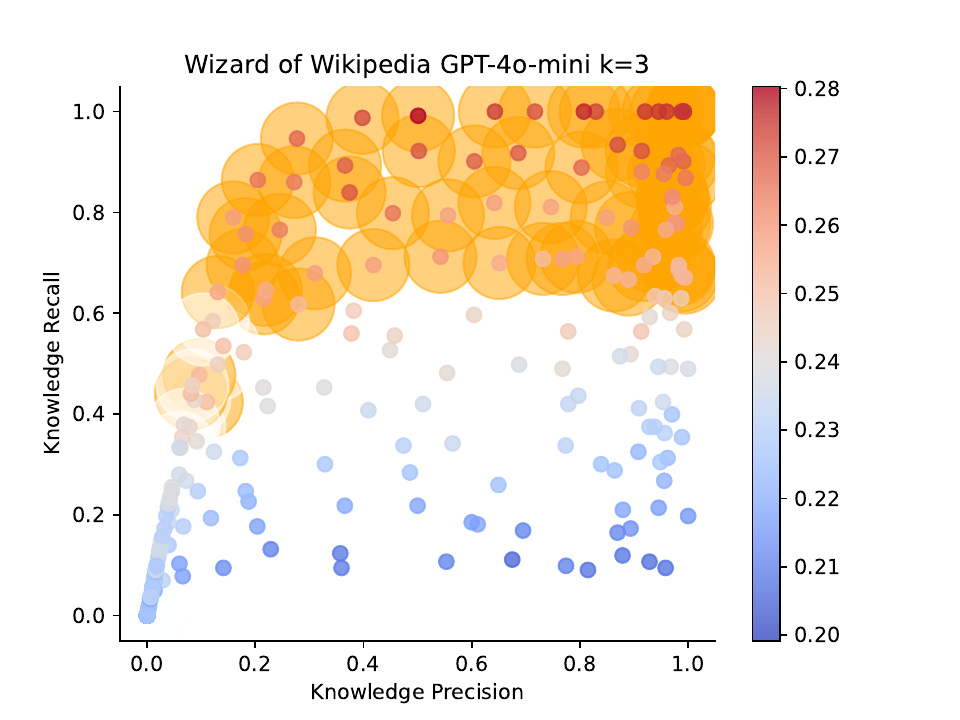}
          \label{fig:wow_gpt_selection_prec_recall_by_area_k=3}
          \vspace{-1.5em}
        \end{subfigure}
        \quad
      \begin{subfigure}[b]{0.48\textwidth}
        \centering
          \includegraphics[width=\textwidth]{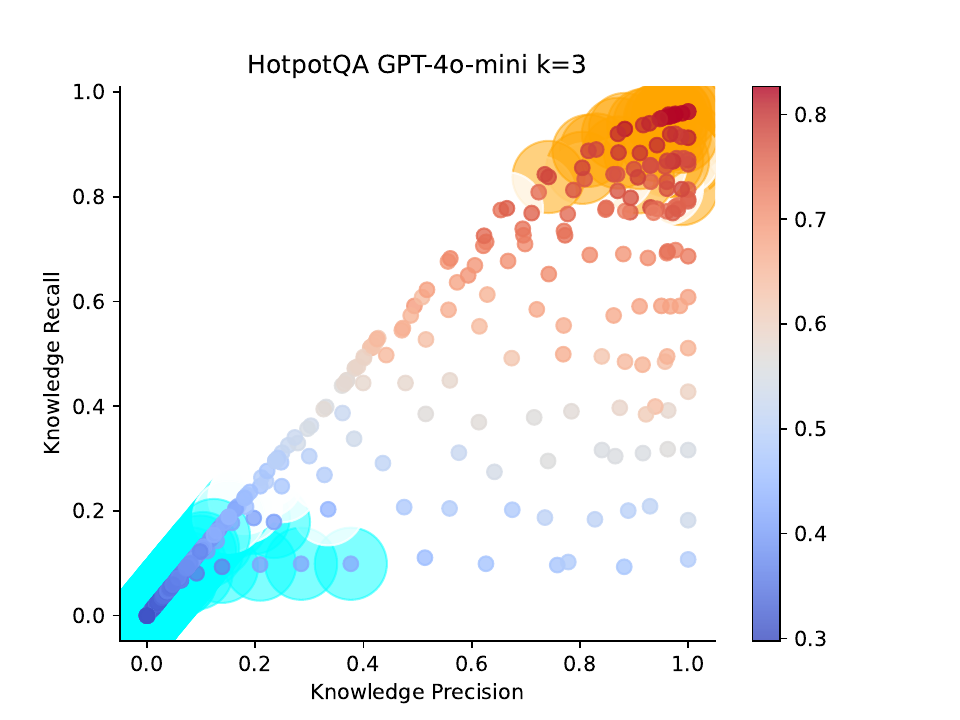}
          \label{fig:hotpot_qa_gpt_selection_prec_recall_by_area_k=3}
          \vspace{-1.5em}
        \end{subfigure}
     \quad
       
     \vspace{-0.5em}
        \caption{Scatter plot of answer F1 versus the knowledge precision for GPT-4o-mini on Wizard of Wikipedia (left) and HotpotQA (right) by limiting the number of knowledge sentences up to k=3. The dots highlighted in orange indicate settings outperforming the ``full knowledge'' setting (without constraining the number of sentences for fair comparison with other figures), while those highlighted in cyan indicate settings underperforming the ``no knowledge" setting. \textit{Each figure is a meta-experiment, and each data point corresponds to a full experiment on the entire sampled dataset.}}
        \label{fig:gpt_selection_prec_recall_by_area_k=3}
\end{figure*}

As Figure \ref{fig:knowledge_selection_precision_recall_vs_k} shows, the length of the knowledge input mostly correlates with knowledge precision. As Figure \ref{fig:gpt_selection_prec_recall_by_area_k=3} shows, constraining the knowledge input size pushes the points in the upper-left corner of the plots in Figure \ref{fig:selection_prec_recall_vs_answer_f1} into the rest of the space while keeping the overall trend of the points the same. Our observations in Section \ref{sec:RagSimulation_results} hold even with length-constrained knowledge selection: a sufficiently high knowledge recall score is a pre-requisite for a knowledge selector outperform the ``full knowledge'' setting. 

\begin{table}[t]
\small
\begin{center}
\setlength{\tabcolsep}{5pt} 
    \begin{tabular}{p{0.9\linewidth} }
    \hline
    \textbf{Prompt} \\ \hline
The following is the conversation between the "Wizard", a knowledgable speaker who can access to Wikipedia knowledge sentences to chat to with the "Apprentice", who does not have access to Wikipedia. \\
\\
The conversation is about ``\{\{persona\}\}''.\\

\{\% if history \%\} \\
Here is the conversation history:\\
\{\% for turn in history \%\}\\
\{\{turn.speaker\}\}: \{\{turn.text\}\}\\
\{\% endfor \%\}\\
\{\% endif \%\}\\
\\
\{\% if context \%\}\\
Here are some retrieved Wikipedia knowledge for the Wizard. The Wizard can choose any subset of the following knowledge. It's also allowed to not choosing any of them.\\
\\
\{\% for evidence in context \%\}\\
Title: \{\{ evidence.title \}\}\\
Sentences:\\
\{\% for sentence in evidence.sentences \%\}\\
- \{\{ sentence \}\}\\
\{\% endfor \%\}\\
\\
\{\% endfor \%\}\\
\{\% endif \%\}\\
\\
Given the knowledge above, make a very brief, such as one sentence, natural response for the Wizard. \\
Not all information in the chosen knowledge has to be used in the response. Do not include the speaker's name in the response.\\
The Wizard's response is:\\

   \hline
    \end{tabular}
    \caption{Jinja2 prompt template for Wizard of Wikipedia.} \label{tab:WoW_prompt}
\end{center}
\end{table}

\begin{table}[t]
\small
\begin{center}
\setlength{\tabcolsep}{5pt} 
    \begin{tabular}{p{0.9\linewidth} }
    \hline
    \textbf{Prompt} \\ \hline
    Answer this question from HotpotQA with a response that is as short as possible, e.g. one word: \\
    \{\{ question \}\}\\
    
    \{\% if context \%\}\\
    Use the following support evidence to answer: \\
    \{\% for evidence in context \%\}\\
    Title: \{\{ evidence.title \}\}\\
    Sentences:\\
    \{\% for sentence in evidence.sentences \%\}\\
    - {{ sentence }}\\
    \{\% endfor \%\}\\
    \\
    \{\% endfor \%\}\\
    \{\% endif \%\}\\
   \hline
    \end{tabular}
    \caption{Jinja2 prompt template for HotpotQA.} \label{tab:hotpotQA_prompt}
\end{center}
\end{table}


\end{document}